\documentclass{article}

    \usepackage[preprint]{neurips_data_2024}

\usepackage[utf8]{inputenc} 
\usepackage[T1]{fontenc}    
\usepackage{hyperref}       
\usepackage{url}            
\usepackage{booktabs}       
\usepackage{amsfonts}       
\usepackage{nicefrac}       
\usepackage{microtype}      
\usepackage{xcolor}         

\setcitestyle{square}
\usepackage{graphicx}
\usepackage{booktabs}

\usepackage{multirow}
\usepackage{siunitx}
\usepackage{arydshln}
\usepackage{pifont}
\usepackage{subcaption}
\usepackage{enumitem}
\usepackage{wrapfig}

\definecolor{my_green}{RGB}{51,102,0}
\definecolor{my_red}{RGB}{204, 0, 0}
\newcommand{\cmark}{\textcolor{my_green}{\ding{51}}}

\newcommand{\MyParagraph}[1]{\vspace{0mm} \noindent \textit{#1} \hspace{0mm}}
\usepackage{xspace}
\newcommand{\benchmarkname}{MMWorld\xspace}

\title{
\benchmarkname: Towards Multi-discipline Multi-faceted World Model Evaluation in Videos
}

\author{Xuehai He$^{1}$ \quad Weixi Feng\thanks{Equal Contribution}~$^{\ 2}$ \quad  Kaizhi Zheng$^{*1}$ \quad Yujie Lu$^{*2}$ \quad Wanrong Zhu$^{*2}$ \quad Jiachen Li$^{*2}$ \\   
\textbf{Yue Fan$^{*1}$ \quad Jianfeng Wang$^{3}$ \quad Linjie Li$^{3}$ \quad Zhengyuan Yang$^{3}$ \quad Kevin Lin$^{3}$} \\ \textbf{William Yang Wang$^{2}$ \quad Lijuan Wang$^{3}$ \quad Xin Eric Wang$^{1}$}
\\ 
  $^1$UC Santa Cruz \quad $^2$UC Santa Barbara \quad $^3$Microsoft \\
  \texttt{\{xhe89,xwang366\}@ucsc.edu} \\
  \url{https://mmworld-bench.github.io/}
}
  
\begin{document}

\maketitle

\begin{figure}[h]
  \centering
  \includegraphics[width=\textwidth]{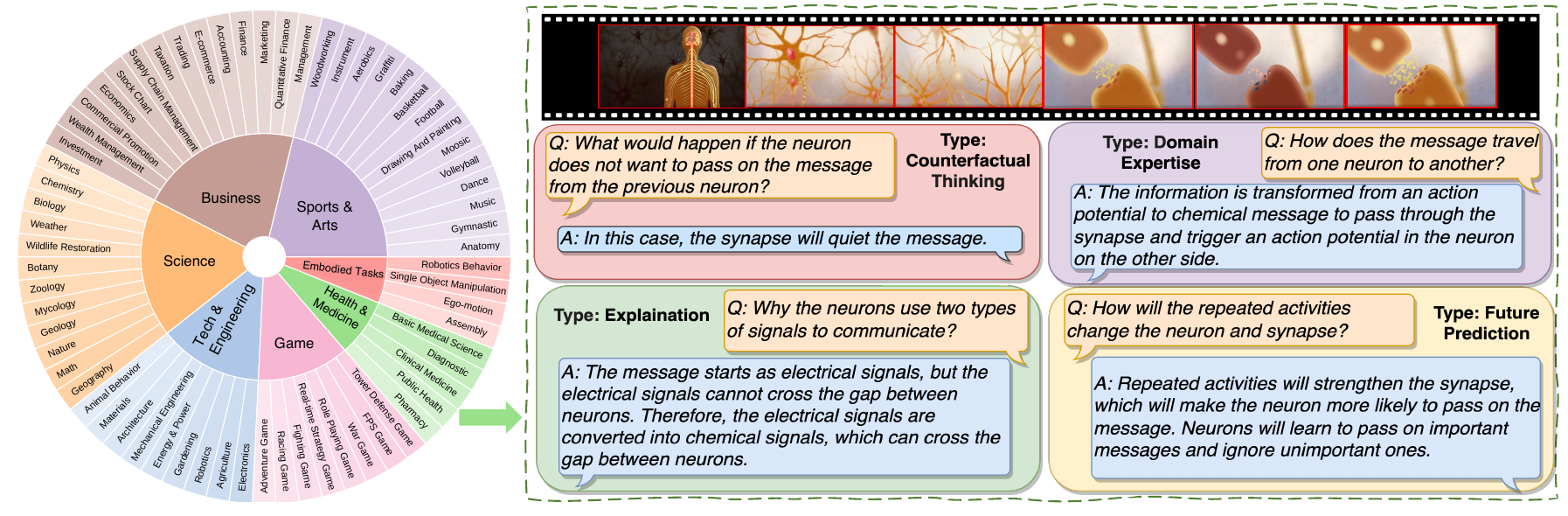}
  \caption{\benchmarkname covers seven broad disciplines and  69 subdisciplines, 
  focusing on the evaluation of multi-faceted reasoning beyond perception (e.g., explanation, counterfactual thinking, future prediction, domain expertise). On the right is a video sample from the Health \& Medicine discipline. }
  \label{fig:overview}
\end{figure}

\begin{abstract}
Multimodal Language Language Models (MLLMs) demonstrate the emerging abilities of "world models"---interpreting and reasoning about complex real-world dynamics. 
To assess these abilities, we posit videos are the ideal medium, as they encapsulate rich representations of real-world dynamics and causalities.
To this end, we introduce \benchmarkname, a new benchmark for multi-discipline, multi-faceted multimodal video understanding.
\benchmarkname distinguishes itself from previous video understanding benchmarks with two unique advantages: (1) \textbf{multi-discipline}, covering various disciplines that often require domain expertise for comprehensive understanding; (2) \textbf{multi-faceted reasoning}, including explanation, counterfactual thinking, future prediction, etc.
\benchmarkname consists of a human-annotated dataset to evaluate MLLMs with questions about the whole videos and a synthetic dataset to analyze MLLMs within a single modality of perception. 
Together, \benchmarkname encompasses 1,910 videos across seven broad disciplines and 69 subdisciplines, complete with 6,627 question-answer pairs and associated captions. 
The evaluation includes 2 proprietary and 10 open-source MLLMs, which struggle on \benchmarkname (e.g., GPT-4V performs the best with only 52.3\% accuracy), showing large room for improvement. Further ablation studies reveal other interesting findings such as models' different skill sets from humans. We hope \benchmarkname can serve as an essential step towards world model evaluation in videos. 
\end{abstract}

\section{Introduction}
\label{sec:intro}

Foundation models, such as Large Language Models (LLMs)~\citep{openai2023gpt4, touvron2023llama, jiang2023mistral,anil2023palm} and Multimodal LLMs (MLLMs)~\citep{gpt4-v,team2023gemini,videollava,li2023videochat,Maaz2023VideoChatGPT,chen2023minigptv2}, have demonstrated remarkable abilities in text and image domains, igniting debates about their potential pathways to Artificial General Intelligence (AGI). This raises a critical question: how well do these models understand the dynamics of the real world? Are they equipped with an inherent World Model~\citep{lecun2022path,worldknowledge,worldmodels,pandora} that can understand and reason about the underlying principles and causalities of the dynamic, multimodal world?

Videos, with their rich, dynamic portrayal of the real world, are ideally suited for evaluating the "world modeling" capabilities of MLLMs.
Existing video understanding benchmarks~\citep{mvbench,videobench,perceptiontest,mvbench}, however, fall short in two key perspectives for such evaluations. 
First, as LeCun et al.~\citep{lecun2022path} discussed, the world model should be able to \emph{(1) estimate missing information about the state of the world not provided by perception, and (2) predict plausible future states of the world}. Evaluation of such capabilities requires \textbf{multi-faceted reasoning} beyond perception level, including explaining the video dynamics, counterfactual thinking of alternative consequences, and predicting future activities within videos.
Moreover, the \textbf{multi-discipline} nature of the multimodal world necessitates a grasp of diverse fundamental principles---ranging from physics and chemistry to engineering and business. 
Hence, domain expertise across a variety of disciplines is imperative for a thorough evaluation of a model’s world understanding towards AGI~\citep{morris2023agi,yue2023mmmu}.

Therefore, we introduce \benchmarkname, a multi-discipline multi-faceted multimodal video understanding benchmark to comprehensively evaluate MLLMs' abilities in reasoning and interpreting real-world dynamics~\footnote{Note that \benchmarkname is not a sufficient testbed for world model evaluation, but we believe overcoming the unique challenges presented in \benchmarkname is essential and necessary towards comprehensive world modeling.}. \benchmarkname encompasses a wide range of disciplines and presents multi-faceted reasoning challenges that demand a combination of visual, auditory, and temporal understanding. 
It consists of 1,910 videos that span seven common disciplines, including \emph{Art \& Sports}, \emph{Business}, \emph{Science}, \emph{Health \& Medicine}, \emph{Embodied Tasks}, \emph{Tech \& Engineering}, and \emph{Games}, and 69 subdisciplines (see Figure~\ref{fig:overview}) such as Robotics, Chemistry, Trading, and Agriculture, thereby fulfilling the objective of breadth in discipline coverage. The dataset includes a total of 1,559 question-answer pairs and video captions annotated and reviewed by humans.
Meanwhile, for multi-faceted reasoning, \benchmarkname mainly contains seven kinds of questions focusing on \emph{explanation} (explaining the phenomenon in videos), \emph{counterfactual thinking} (answering what-if questions), \emph{future prediction} (predicting future events), \emph{domain expertise} (answering domain-specific inquiries), \emph{temporal understanding} (reasoning about temporal information), and etc. A video example with these four questions from the Health \& Medicine discipline is depicted in Figure~\ref{fig:overview}. \benchmarkname comprises two datasets: a human-annotated dataset for evaluating MLLMs on the whole video and a synthetic dataset designed to analyze MLLMs' perception within single visual or audio modalities. We evaluate 12 MLLMs that can handle videos or image sequences on \benchmarkname, including both open-source (e.g., Video-LLaVA-7B~\citep{videollava}) and proprietary models (GPT-4V~\citep{gpt4-v} and Gemini~\citep{team2023gemini}).

We summarized the contributions and key findings as follows:
\begin{itemize} 
\item We introduce \benchmarkname, a new benchmark designed to rigorously evaluate the capabilities of Multimodal Large Language Models (MLLMs) in world modeling through the realm of video understanding. \benchmarkname spans a broad spectrum of disciplines, featuring a rich array of question types for multi-faceted reasoning. 
\item In addition to the human-annotated dataset, we develop an automatic data collection pipeline, streamlining video content selection and question-answer generation, and construct a well-controlled synthetic dataset to analyze MLLMs within single visual or audio modalities.
\item We observe that existing MLLMs still face substantial challenges posed by \benchmarkname. Even the best performer, GPT-4o, can only achieve a 52.30\% overall accuracy, and four MLLMs particularly trained on videos perform worse than random chance.
\item Although there is stll a clear gap between open-source and proprietary models, the best open-source model Video-LLaVA-7B outperforms GPT-4V and Gemini on Embodied Tasks by a large margin and performs similarly on Art \& Sports, where spatiotemporal dynamics play a more crucial role in video understanding. This is further validated with its leading results on the Temporal Understanding question type. 
\item In our study comparing MLLMs with average humans (non-experts), we notice some correlation between question difficulties as perceived by humans and MLLMs. However, MLLMs present different skill sets than humans in that they can answer reasonable amount of difficult questions that humans completely fail but also struggle at easy questions that humans excel at. This indicates different perception, cognition, and reasoning abilities between MLLMs and humans.
\end{itemize}

\begin{table}[t]
\centering
\caption{Comparison between \benchmarkname and previous benchmarks for real-world video understanding on a variety of criteria. Multi-faced include Explanation (\texttt{Explain.}), Counterfactual Thinking (\texttt{Counter.}), Future Prediction (\texttt{Future.}) and Domain Expertise (\texttt{Domain.}) \benchmarkname is the first multi-discipline and multitask video understanding benchmark that covers wider reasoning questions, and also included first-party data annotations.  }
\label{tab:video-datasets}
\setlength{\tabcolsep}{3pt}
\resizebox{\linewidth}{!}{
\begin{tabular}{lccccccc}
\toprule
\multirow{2}{*}[0em]{\textbf{Benchmarks}} 
& \multirow{2}{*}[0em]{ $\begin{array}{l}
    \textbf{Multi-}\\
     \textbf{Discipline} \\
\end{array}$}
& \multirow{2}{*}[0em]{$\begin{array}{l}
    \textbf{Multi-}\\
     \textbf{Task} \\
\end{array}$}
& \multicolumn{4}{c}{\textbf{Multi-Faceted Reasoning}} 
& \multirow{2}{*}[0em]{$\begin{array}{l}
    \textbf{First-Party}\\
     \textbf{Annotation} \\
\end{array}$} \\
\cmidrule{4-7}
&&
& $\begin{array}{c}
    \texttt{Explain.}\\
\end{array}$      
& $\begin{array}{c}
    \texttt{Counter.}\\
\end{array}$
& $\begin{array}{c}
    \texttt{Future.}\\
\end{array}$
& $\begin{array}{c}
    \texttt{Domain.}\\
\end{array}$
&  \\
\midrule
MovieQA~\citep{tapaswi2016movieqa} &  &  &\cmark&&&  & \cmark 
\\TVQA~\citep{lei2018tvqa} &  &  &\cmark&&&  & \cmark \\
ActivityNet-QA~\citep{yu2019activitynet} &  &  & &&&  & \cmark  \\
MSVD-QA~\citep{xu2017video}~\citep{msr-vtt} &  &  &&&&  & \cmark  \\
MSRVTT-QA~\citep{msr-vtt} &  &  &&&&  & \cmark  \\
Sports-QA~\citep{sportsqa} &  &  & &\cmark&& \cmark & \cmark  \\
VaTeX~\citep{wang2019vatex} &  & \cmark &  &  &  &  & \cmark  \\
VALUE~\citep{li2021value} &  & \cmark &  &  &  &  &   \\
Video-Bench~\citep{ning2023video} &  & \cmark & &&\cmark& \cmark &   \\
MVBench~\citep{mvbench} &  & \cmark &&\cmark&\cmark&  &  \\
Perception Test~\citep{perceptiontest} &  & \cmark &\cmark&\cmark&\cmark&  &  \\
\benchmarkname (Ours) & \cmark & \cmark & \cmark & \cmark & \cmark & \cmark & \cmark  \\
\bottomrule
\end{tabular}}
\end{table}

\section{Related Work}
\label{sec:related work}
\subsection{Multimodal Large Language Models (MLLMs)}
\paragraph{Emerging MLLMs~}
With recent breakthroughs~\citep{gpt4,google2023bard,touvron2023llama,vicuna2023,touvron2023llama2,bai2023qwen} in Large Language Models (LLMs), several counterparts in the vision-and-language domain have been proposed~\citep{dai2023instructblip,liu2023visual,liu2023improved,li2023otter,zhu2023minigpt,zheng2023minigpt,bai2023qwenvl}, and recently released GPT-4V~\citep{gpt4-v}, followed by Gemini Vision family~\citep{team2023gemini}. 
Many MLLMs have expanded their capabilities beyond handling only text and image inputs. VideoChat~\citep{li2023videochat} leverages the QFormer~\citep{blip2} to map visual representations to LLM~\citep{vicuna2023}, and performs a multi-stage training pipeline. Otter~\citep{li2023otter} proposes to conduct instruction finetuning based on Openflamingo~\citep{awadalla2023openflamingo}. PandaGPT~\citep{su2023pandagpt} employs the ImageBind~\citep{han2023imagebind} as the backbone and finetunes it. mPLUG-Owl~\citep{ye2023mplug} introduces an abstractor module to perform visual and language alignment. VideoLLaMA~\citep{zhang2023videollama} introduces a frame embedding layer and also leverages ImageBind to inject temporal and audio information into the LLM backend. Chat-UniVi~\citep{jin2023chatunivi} uses clustering to do feature fusion. Observing their emerging abilities in multimodal video understanding, we propose~\benchmarkname to evaluate these models' skills in understanding the dynamics of the real world.

\paragraph{Benchmarking MLLMs~}
To evaluate MLLMs, there is a flourishing of analysis \citep{liu2023mitigating,zhang2023gpt4vision,comclip,yujie2024wildvisionarena,fan2024muffin,cui2023holistic, guan2023hallusionbench,yu2023mmvet,fu2023mme} and the establishment of innovative benchmarks such as VisIB-Bench~\citep{bitton2023visit} which evaluates models with real-world instruction-following ability given image inputs, MMMU~\citep{yue2023mmmu} designed to access models on college-level image-question pairs that span among different disciplines, and VIM~\citep{lu2023vim} which challenges the model's visual instruction following capability.
However, these recent analyses and benchmarks only cover the image input, which hinders the evaluation of MLLM's performance as a world model. Recently, video benchmarks such as Perception Test~\citep{perceptiontest} is proposed to focus on perception and skills like memory and abstraction. However, it uses scenarios with a few objects manipulated by a person, which limits the variety of contexts. MVBench~\citep{mvbench} centers on temporal understanding, while \benchmarkname not only includes temporal reasoning but also evaluates other multi-faceted reasoning abilities.


\subsection{Video Understanding Benchmarks}
Previous video benchmarks, as shown in Table~\ref{tab:video-datasets}, focus on video understanding tasks, including activity-focused on web videos~\citep{activityQA}, description-based question answering~\citep{VideoQA}, video completion~\citep{fu2023tvc}, and video infilling~\citep{himakunthala2023lets}. Recently, Video-Bench~\citep{videobench} introduces a benchmark by collecting videos and annotations from multiple existing datasets. LWM~\citep{lwm} collects a large video and language dataset from public books and video datasets and trains a world model that is capable of processing more than millions of tokens. However, modeling
millions of tokens is extremely difficult due to high memory cost, computational complexity, and lack
of suitable datasets. Mementos~\citep{wang2024mementos} builds a benchmark for MLLM reasoning for input image sequences. STAR~\citep{star} builds a benchmark for situated reasoning in real-world videos. CLEVER~\citep{clevrer} builds a benchmark containing videos focusing on objects with simple visual appearance. 
Our contribution, in contrast, presents a new video understanding benchmark designed to evaluate models on several pivotal components crucial for a comprehensive world model. These components encompass interdisciplinary coverage, task diversity, and multifaceted reasoning capabilities—including future prediction, counterfactual thinking, and more—underpinned by original human annotations and integrated domain knowledge.

\section{The \benchmarkname Benchmark}
The \benchmarkname benchmark is built on three key design principles: multi-discipline coverage and multi-faceted reasoning. It spans various disciplines that require domain expertise and incorporates diverse reasoning skills such as explanation, counterfactual thinking, and future prediction. The benchmark consists of two parts: a human-annotated dataset and a synthetic dataset. The human-annotated dataset serves as the main test bed to evaluate MLLMs from multiple perspectives. The synthetic dataset contains two subsets, focusing on evaluating MLLMs' perception behavior from both visual signals and audio inputs, respectively.

\subsection{Manual Data Collection}
\label{sec:data_collection}
We collect videos from YouTube with the Creative Licence in seven disciplines:
Art $\&$ Sports (18.5\%), Business (12.0\%), Science (20.4\%), Health $\&$ Medicine (12.0\%), Embodied Tasks (12.0\%\%), Tech $\&$ Engineering (12.9\%), and Game (12.2\%). For Art $\&$ Sports, 29 videos are collected from the SportsQA dataset~\citep{sportsqa}. And for Embodied Tasks, 24 videos are sourced from IKEA Assembly~\citep{ben2021ikea}, RT-1~\citep{rt-1}, and Ego4D~\citep{grauman2022ego4d} datasets to increase video diversity.

Our manual benchmark collection takes two
stages. In the first stage, we conduct a detailed examination of each of the seven primary disciplines to identify a comprehensive range of subdisciplines for inclusion in our benchmark. 
Our selection of videos is driven by three key principles:
\begin{itemize}
    \item 
The \textbf{first principle}, \textbf{multi-discipline} coverage, emphasizes the requirement for domain knowledge—selecting videos that inherently demand an understanding of specialized content across various disciplines. 
  \item 
  The \textbf{second principle}, \textbf{multi-faceted} annotation, involves collecting videos that enable the creation of question-answer pairs from multiple perspectives to evaluate world model properties comprehensively. 
    \item  The \textbf{third principle}, \textbf{temporal information}, prioritizes the inclusion of videos that provide meaningful content over time, as understanding temporal information is crucial for grasping world dynamics. This allows models to engage in temporal reasoning.
Therefore, answering questions in our dataset requires implicit temporal reasoning, e.g., the model needs to understand temporal information to explain ``why does the robot need to do the step shown in the video''. We also design a ``temporal understanding'' question type to explicitly test models' ability to reason about temporal information (examples can be found in Section F in the Appendix).
\end{itemize}

\begin{figure}[tb]
  \centering
    \includegraphics[width=\textwidth]{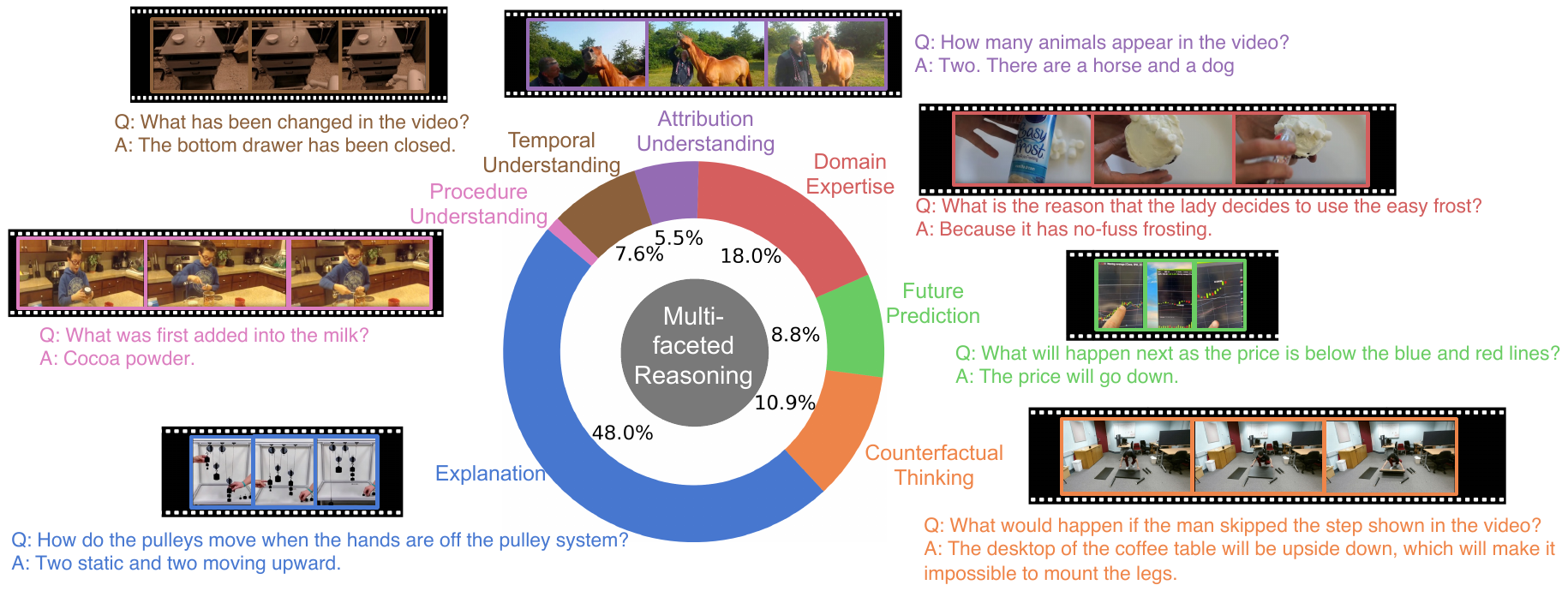}
    \caption{The questions in \benchmarkname primarily evaluate seven understanding and reasoning abilities of models. We give one example for each category. }
  \label{fig:questions_per_type}
\end{figure}

During the second stage, our team embark on the task of question annotation. We craft questions that primarily test seven aspects of multimodal video understanding also from the perspective of \textbf{multi-faceted reasoning}: 1) Explanation: Questions ask the model to elucidate the underlying logic or purpose within the video;
2) Counterfactual Thinking: Tests the model's ability to hypothesize and consider alternative outcomes;
3) Future Prediction: Aims to predict future events based on the current scenario, challenging the model’s foresight;
4) Domain Expertise: Evaluates the model's depth of knowledge in specific fields, such as how to assemble a coffee table;
5) Temporal Understanding: Assesses the model's capability to reason about temporal sequences and dynamics;
6) Attribution Understanding: These questions focus on identifying cause-and-effect relationships within the video, including tasks like counting;
7) Procedure Understanding: Tests the model's ability to comprehend and explain procedural tasks shown in the video. The detailed distribution and examples are shown in Figure~\ref{fig:questions_per_type}.

\subsection{Automated Data Collection}
\label{automatic_data}
Understanding real-world dynamics requires models to process both audio and visual modalities. To evaluate MLLMs' perception abilities in these modalities, we designed an automated data collection pipeline. This pipeline collects targeted videos and generates QA pairs based on either audio or visual information, ensuring the model's capabilities are assessed independently for each modality. By using information from a single modality to generate QA pairs, our pipeline ensures that the synthetic data remains unbiased regarding input modality.

\begin{figure}[tb]
  \centering
  \includegraphics[width=\textwidth]{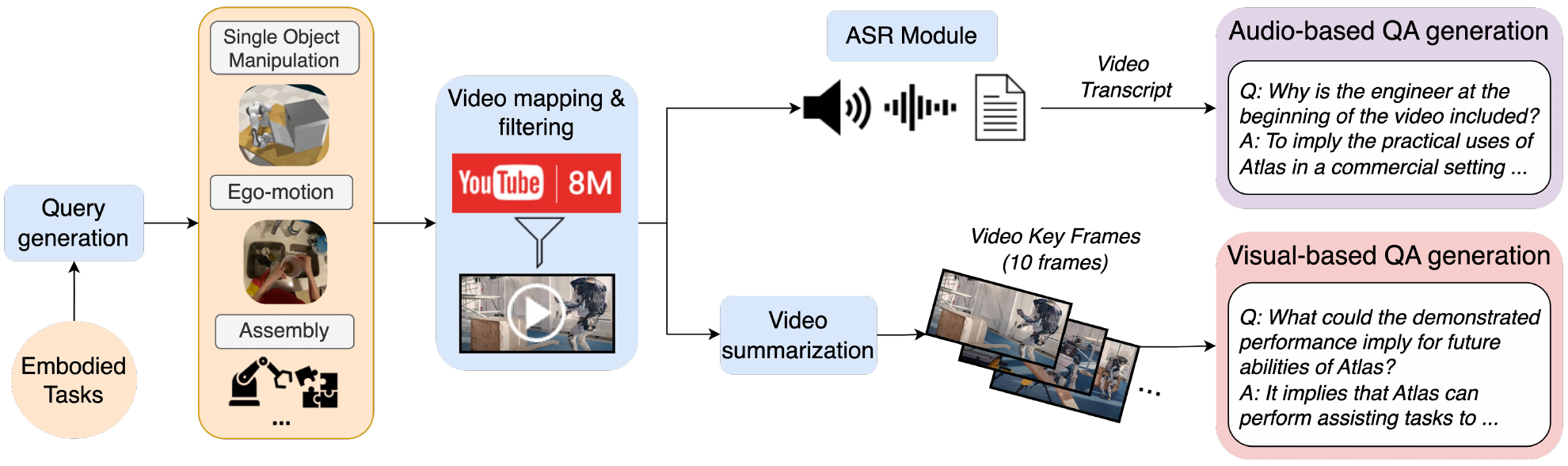}
  \caption{Schematic diagram of the synthetic data generation pipeline in \benchmarkname. It starts with generating subdiscipline-specific queries, followed by video retrieval from YouTube-8M~\citep{youtube8m} and YouTube. Keyframes are extracted for visual-based QA generation, and videos are transcribed using an ASR module for audio-based QA generation. 
  }
  \label{fig:example}
\end{figure}

The synthetic data generation pipeline is illustrated in Figure~\ref{fig:example}. We employ a systematic approach to gather videos with Creative Commons licenses from YouTube and the extensive YouTube-8M dataset~\citep{youtube8m}. This method ensures a diverse and comprehensive collection of video data, which is important for the robust evaluation of multimodal video understanding models.

\paragraph{Video Collection and Processing~}
We start with the video~\textit{Query Generator}. We start with the same seven disciplines as the manually collected dataset. For each discipline, a set of subdisciplines is defined to encapsulate a wide spectrum of topics, ensuring a diverse and comprehensive dataset.
Once the queries are generated, the \textit{Video Mapping and Filtering} step is initiated. We perform mapping of videos to YouTube-8M and online videos, constrained by a strict time limit of two minutes per query, keeping only the most pertinent videos that satisfy the predefined criteria. Simultaneously, the works in conjunction with the video transcripts to extract key terms and concepts. This iterative process refines the search parameters and enhances the semantic richness of the dataset by identifying and encoding the salient themes present in the videos. The \textit{Video Summarization} module utilizes Query-focused video summarization techniques based on Katna\footnote{
\url{https://github.com/keplerlab/katna}} and UniVTG~\citep{univtg}. This module selects ten representative frames from each video, distilling the essence of the content while preserving the narrative context. This summarization facilitates efficient storage and quicker processing times, which are crucial for large-scale analysis.

\paragraph{QA Generation~}
The final stage in our pipeline is the \textit{QA / Caption Generation} module, where we leverage the capabilities of GPT-4V to generate accurate and contextually relevant questions and answers, as well as captions, based on the video frames and transcripts. This step not only provides rich annotations for each video but also equips the dataset with a multimodal dimension that supports various downstream tasks such as video QA, captioning, and more.

\paragraph{Quality of the Synthetic Dataset~}
Human evaluators were engaged to ascertain the reasonableness of automatically generated questions and answers, ensuring that the synthetic dataset maintains a high standard of quality and relevance. The findings from this human evaluation phase are detailed in Section D of the Appendix, offering insights into the dataset's efficacy and the realism of its constructed queries and responses.

\begin{table}{t}
\caption{Key Statistics of the \benchmarkname Benchmark. The main subset is the human-annotated subset. Synthetic Subset I contains generated QA pairs focused exclusively on the audio content, while Synthetic Subset II contains QA pairs focused exclusively on the visual content of the video.}
\centering
\begin{tabular}{lrrrr}
\toprule
\multicolumn{1}{l}{\textbf{Statistics}} & \multicolumn{1}{l}{\textbf{Main Subset}} & \multicolumn{1}{l}{\textbf{Synthetic I}} & \multicolumn{1}{l}{\textbf{Synthetic II}}  \\
\midrule
\#Discipline/\#Subdiscipline & 7/61 & 7/51 & 7/54  \\
\#<Video-QA> & <417-1,559> & <746-2,969> & <747-2,099>  \\
Avg Video Lengths (s) & 102.3 & 103.4 & 115.8 \\
\hdashline
Avg \#Questions per Video & 4.05 & 3.98 & 2.81  \\
Avg \#Options & 3.90 & 4.00 & 4.00  \\
Avg Question Length & 11.39 & 15.12 & 17.56 \\
Avg Option Length & 7.27 & 6.01 & 5.19  \\
Avg Answer Length & 6.42 & 6.71 & 5.67  \\
Avg Caption Length & 27.00 & 71.87 & 82.33  \\
\bottomrule
\end{tabular}
\label{tab:merged_benchmark_stats_total}
\end{table}
Finally, the statistics of automated curated data, which is used for the ablation study, are shown in Table~\ref{tab:merged_benchmark_stats_total}.
The taxonomy of our dataset is shown in Figure~\ref{fig:overview}. We note that only a portion of the subdisciplines are shown due to space concerns. Please refer to the Appendix for full information.

\section{Experiments}
\subsection{Experimental Settings}
\label{sec:implement}
In our study, we compare MLLM's performance on the \benchmarkname benchmark, including GPT-4V~\citep{gpt4-v}, Gemini Pro~\citep{team2023gemini}, Video-Chat~\citep{li2023videochat}, Video-LLaMA~\citep{zhang2023videollama}, ChatUnivi~\citep{jin2023chatunivi}, mPLUG-Owl~\citep{ye2023mplug}, Otter~\citep{li2023otter}, ImageBind-LLM~\citep{han2023imagebind}, PandaGPT~\citep{su2023pandagpt}, LWM~\citep{lwm}, 
and X-Instruct-BLIP~\citep{panagopoulou2023xinstructblip}. For both Gemini Pro and GPT-4V, we adhere to the default settings provided by their official APIs. They both take ten image frames extracted from the video content as the input. The Gemini Pro is set to process visual input and configured with safety settings to filter a range of harmful content. The configuration thresholds are set to `BLOCK\_NONE'. For PandaGPT, we set `top\_p' to 0.7 and `temperature' to 0.5. For VideoChat, we set `max\_frames' to 100. For X-Instruct-BLIP, the model is implemented using four image frames. We use GPT-4-32K as the judge for judging whether the model answer is correct when it can not mapped to the option letter using the rule-based method. For others, we all use the default setting. All inferences are run on a NVIDIA A6000 workstation. The detailed implementation is given in the Appendix.

\begin{table}[t]
\centering
\caption{MLLM accuracy across diverse disciplines (averaging over three runs). 
GPT-4V and Gemini Pro lead at most disciplines and achieve the best overall accuracy. The best open-source model Video-LLaVA-7B outperforms them on Embodied Tasks and perform similarly on Art \& Sports.}
\label{maineval}
\setlength{\tabcolsep}{3pt}
\resizebox{\linewidth}{!}{
\begin{tabular}{llllllllll}
\toprule
\multirow{2}{*}{\textbf{Model}}        
& \textbf{Art\& } & \multirow{2}{*}{\textbf{Business}} & \multirow{2}{*}{\textbf{Science}} & \textbf{Health\&} & \textbf{Embodied } & \textbf{Tech\& } & \multirow{2}{*}{\textbf{Game}} & \multirow{2}{*}{\textbf{Average}} \\
& \textbf{Sports}& & &  \textbf{Medicine}& \textbf{Tasks}& \textbf{Engineering}& \\
\midrule
Random Choice&25.03 & 25.09 & 26.44 & 25.00 & 26.48 & 30.92 & 25.23 & 26.31 \\ 
\midrule
\multicolumn{9}{c}{\textit{Proprietary MLLMs}} \\
\midrule
GPT-4o~\citep{gpt4o} & \underline{47.87} \tiny{$\pm$1.47} & \textbf{91.14} \tiny{$\pm$0.87} & \textbf{73.78} \tiny{$\pm$2.88} & \textbf{83.33} \tiny{$\pm$1.47} & \underline{62.94} \tiny{$\pm$3.47} & \textbf{75.53} \tiny{$\pm$2.61} & \textbf{80.32} \tiny{$\pm$2.05} & \textbf{62.54} \tiny{$\pm$0.79} \\
Claude-3.5-Sonnet~\citep{claude} & \textbf{54.58} \tiny{$\pm$0.45} & 63.87 \tiny{$\pm$0.40} & 59.85 \tiny{$\pm$1.28} & 54.51 \tiny{$\pm$1.28} & 30.99 \tiny{$\pm$0.40} & 58.87 \tiny{$\pm$0.61} & 59.44 \tiny{$\pm$0.68} & \underline{54.54 \tiny{$\pm$0.29}} \\
GPT-4V~\citep{gpt4-v}         & 36.17  \tiny{$\pm$0.58  }& \underline{81.59}  \tiny{$\pm$1.74  }& \underline{66.52}  \tiny{$\pm$1.86  }& 73.61  \tiny{$\pm$0.49  }& 55.48  \tiny{$\pm$2.70  }& 61.35  \tiny{$\pm$1.00  }& \underline{73.49}  \tiny{$\pm$1.97  }& 52.30  \tiny{$\pm$0.49     }        \\
Gemini Pro~\citep{team2023gemini}          & 37.12  \tiny{$\pm$2.68  }& 76.69  \tiny{$\pm$2.16  }& 62.81  \tiny{$\pm$1.83  }& \underline{76.74}  \tiny{$\pm$1.30  }& 43.59  \tiny{$\pm$0.33  }& \underline{69.86}  \tiny{$\pm$2.01  }& 66.27  \tiny{$\pm$2.60  }& 51.02  \tiny{$\pm$1.35 } \\
\midrule 
\multicolumn{9}{c}{\textit{Open-source MLLMs}}  \\
\midrule
Video-LLaVA-7B~\citep{videollava}      & 35.91  \tiny{$\pm$0.96  }& 51.28  \tiny{$\pm$0.87  }& 56.30  \tiny{$\pm$0.76  }& 32.64  \tiny{$\pm$0.49  }& \textbf{63.17}  \tiny{$\pm$1.44  }& 58.16  \tiny{$\pm$1.00  }& 49.00  \tiny{$\pm$3.16  }& 44.60  \tiny{$\pm$0.58   }       \\
Video-Chat-7B~\citep{li2023videochat} & 39.53  \tiny{$\pm$0.06  }& 51.05  \tiny{$\pm$0.00  }& 30.81  \tiny{$\pm$0.21  }& 46.18  \tiny{$\pm$0.49  }& 40.56  \tiny{$\pm$0.57  }& 39.36  \tiny{$\pm$0.00  }& 44.98  \tiny{$\pm$0.57  }& 40.11  \tiny{$\pm$0.06 } \\
ChatUnivi-7B~\citep{jin2023chatunivi} & 24.47  \tiny{$\pm$0.49  }& 60.84  \tiny{$\pm$1.51  }& 52.00  \tiny{$\pm$0.73  }& 61.11  \tiny{$\pm$1.96  }& 46.15  \tiny{$\pm$2.06  }& 56.74  \tiny{$\pm$1.33  }& 52.61  \tiny{$\pm$2.84  }& 39.47  \tiny{$\pm$0.42 } \\
mPLUG-Owl-7B ~\citep{ye2023mplug} & 29.16  \tiny{$\pm$1.62  }& 64.10  \tiny{$\pm$1.84  }& 47.41  \tiny{$\pm$3.29  }& 60.07  \tiny{$\pm$1.30  }& 23.78  \tiny{$\pm$3.47  }& 41.84  \tiny{$\pm$5.09  }& 62.25  \tiny{$\pm$3.16  }& 38.94  \tiny{$\pm$1.52  }
\\
PandaGPT-7B~\citep{su2023pandagpt} & 25.33  \tiny{$\pm$0.54  }& 42.66  \tiny{$\pm$3.02  }& 39.41  \tiny{$\pm$2.67  }& 38.54  \tiny{$\pm$3.07  }& 35.43  \tiny{$\pm$0.87  }& 41.84  \tiny{$\pm$2.79  }& 40.16  \tiny{$\pm$4.65  }& 32.48  \tiny{$\pm$0.45 } \\
ImageBind-LLM-7B~\citep{han2023imagebind} & 24.82  \tiny{$\pm$0.16  }& 42.66  \tiny{$\pm$0.99  }& 32.15  \tiny{$\pm$1.11  }& 30.21  \tiny{$\pm$1.47  }& 46.85  \tiny{$\pm$1.14  }& 41.49  \tiny{$\pm$1.50  }& 41.37  \tiny{$\pm$0.57  }& 31.75  \tiny{$\pm$0.14 } \\
X-Instruct-BLIP-7B~\citep{panagopoulou2023xinstructblip} & 21.08  \tiny{$\pm$0.27  }& 15.85  \tiny{$\pm$0.87  }& 22.52  \tiny{$\pm$1.11  }& 28.47  \tiny{$\pm$0.49  }& 18.41  \tiny{$\pm$1.44  }& 22.34  \tiny{$\pm$0.87  }& 26.10  \tiny{$\pm$0.57  }& 21.36  \tiny{$\pm$0.18 } \\
LWM-1M-JAX~\citep{lwm} & 12.04  \tiny{$\pm$0.53  }& 17.48  \tiny{$\pm$0.57  }& 15.41  \tiny{$\pm$0.91  }& 20.49  \tiny{$\pm$0.98  }& 25.87  \tiny{$\pm$1.98  }& 21.99  \tiny{$\pm$2.19  }& 11.65  \tiny{$\pm$3.01  }& 15.39  \tiny{$\pm$0.32 } \\
Otter-7B~\citep{li2023otter} & 17.12  \tiny{$\pm$1.17  }& 18.65  \tiny{$\pm$0.87  }& ~~9.33  \tiny{$\pm$0.36  }& ~~6.94  \tiny{$\pm$0.98  }& 13.29  \tiny{$\pm$1.51  }& 15.96  \tiny{$\pm$1.74  }& 15.26  \tiny{$\pm$0.57  }& 14.99  \tiny{$\pm$0.77}  \\
Video-LLaMA-2-13B~\citep{zhang2023videollama} & ~~6.15  \tiny{$\pm$0.44  }& 21.21  \tiny{$\pm$0.66  }& 22.22  \tiny{$\pm$1.45  }& 31.25  \tiny{$\pm$1.70  }& 15.38  \tiny{$\pm$1.14  }& 19.15  \tiny{$\pm$1.74  }& 24.90  \tiny{$\pm$5.93  }& 14.03  \tiny{$\pm$0.29 } \\
 \bottomrule
\end{tabular}}
\end{table}

\subsection{Evaluation}
Our dataset includes multiple-choice questions and captions corresponding to each video, enabling tasks such as video question answering and video captioning. We focus on video question answering by evaluating a model’s performance based on its accuracy in selecting the correct answer from the provided options. One challenge lies in reliably parsing the model’s response to map it to one of the predefined choices. To address this, we employ two mapping strategies. We employ two mapping strategies. The first method employs automated scripts to parse the models' predictions and compare the parsed results with the ground truth, similar to the approach used in~\citep{yue2023mmmu}. The second method involves models freely generating answers, which are then evaluated by GPT-4. Given the question, correct answer, and model's prediction, GPT-4 returns a True or False judgment. This approach is based on recent works in model evaluation~\citep{Maaz2023VideoChatGPT, hsu2023gpt,hackl2023gpt,liu2023gpteval}. We validated this method with human evaluators, showing an error rate of 4.76\% across 189 examples, confirming the effectiveness of GPT-4 as an evaluator. Detailed results for human evaluation and for these two different strategies are provided in Appendix B. In the main paper, all results are evaluated using the second approach.

\subsection{Main Evaluation Results}
\label{sec:results}
We show in Table~\ref{maineval} the main evaluation results of different MLLMs. Among these, GPT-4V emerges as the top performer, closely followed by Gemini Pro. Video-LLaVA also demonstrates strong results, primarily due to the extensive training data which consists of 558K LAION-CCSBU image-text pairs and 702K video-text pairs from WebVid~\citep{webvid}. For instruction tuning, datasets were gathered from two sources: a 665K image-text instruction dataset from LLaVA v1.5 and a 100K video-text instruction dataset from Video-ChatGPT~\citep{Maaz2023VideoChatGPT}. This superior performance may also be attributed to Video-LLaVA’s adoption of CLIP ViT-L/14 trained in LanguageBind~\citep{videollava} as its vision model and the inclusion of a large volume of image-video-text pairings within the
training data. On the other hand, models like Otter and LWM perform poorly across most disciplines, possibly due to their weaker backbone and architecture used. Otter uses the LLaMA-7B language encoder and a CLIP ViT-L/14 vision encoder, both of which are frozen, with only the Perceiver resampler module fine-tuned, which may contribute to its lower performance.
Additionally, some MLLMs perform even worse than random, highlighting the challenging nature of \benchmarkname.

\begin{figure}[t]
  \centering
  \includegraphics[width=0.8\textwidth]{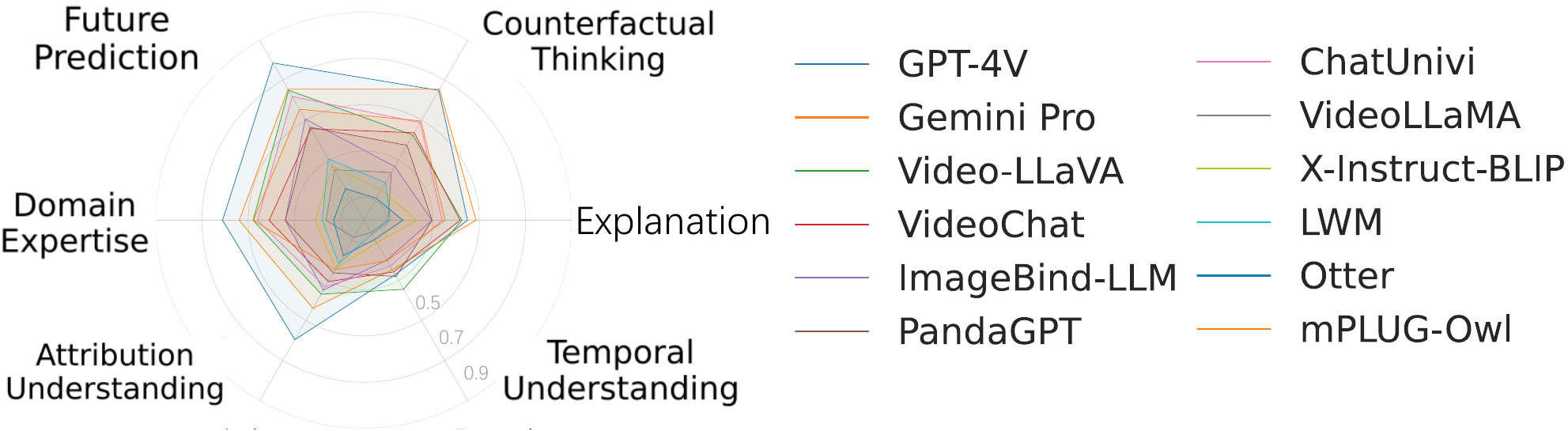}
  \caption{Results of different MLLMs on multi-faceted reasoning. The detailed performance numbers can be found in the Appendix. }
  \label{subdomain_performance}
\end{figure}

\subsection{Study on Multi-faceted Reasoning on \benchmarkname}
Figure~\ref{subdomain_performance} illustrates the multi-faceted reasoning performance for each MLLM. GPT-4V emerges as the strongest model across Future Prediction, Domain Expertise, and Attribution Understanding. Closed-source models like GPT-4V and Gemini Pro perform similarly on counterfactual thinking and outperform all others. However, for temporal understanding, Video-LLaVA performs the best. This may be due to its extensive training on large amounts of video-language data, which enhances its spatio-temporal reasoning abilities. This can be also observed in its high scores on the Art \& Sports and Embodied Tasks, which involve dense spatio-temporal information, as shown in Table~\ref{maineval}. Video-LLaVA's performance is comparable to GPT-4V and Gemini on explanation tasks, likely because of its two-stage training process and exposure to a large amount of instruction-tuning data in the second stage, which includes similar instructions.

\subsection{Study on MLLM Performance at Different Difficulty Levels for Average Humans}

\begin{figure}[tb]
  \centering
  \begin{subfigure}[b]{0.5\textwidth}
    \centering
    \includegraphics[width=\textwidth]{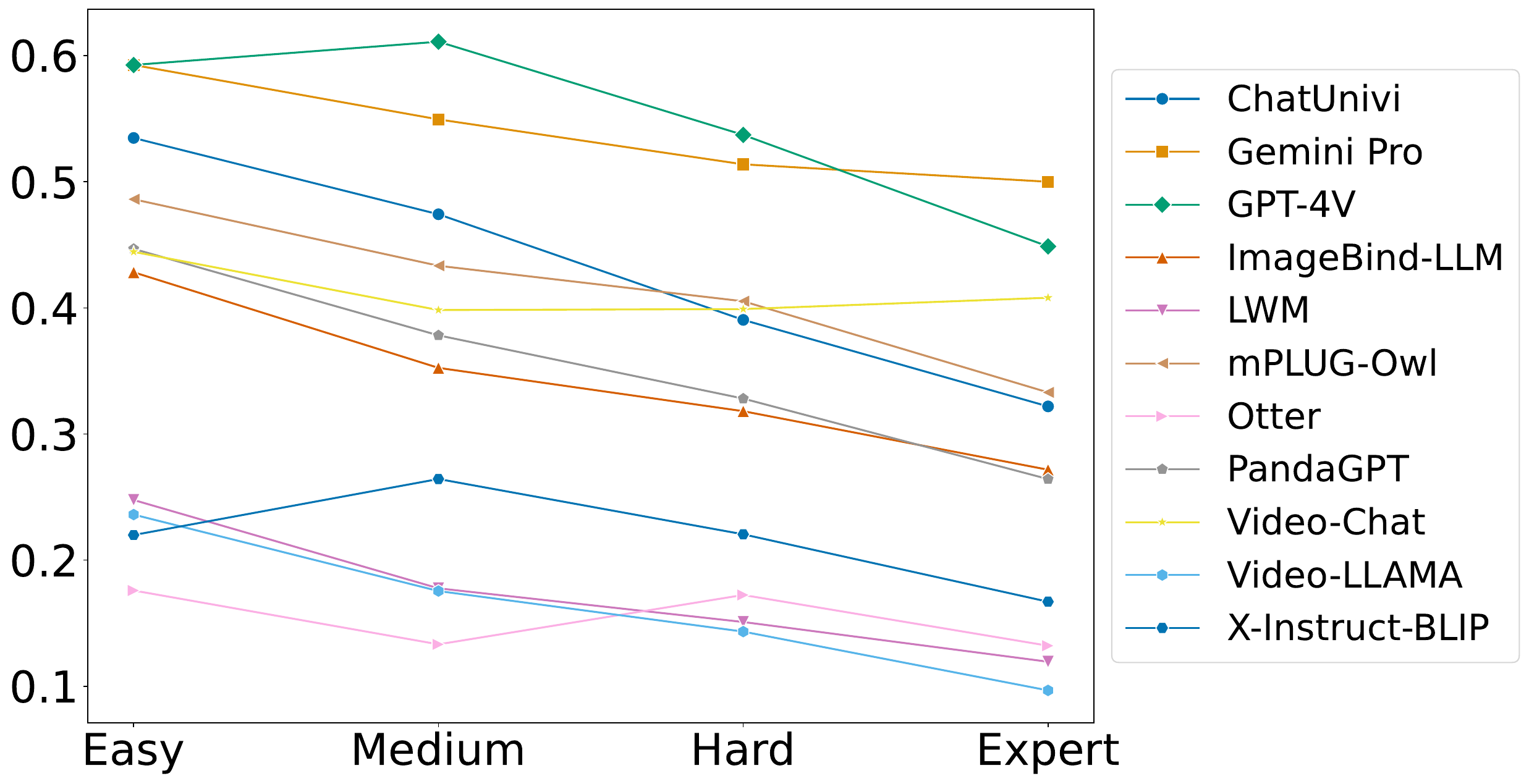}
    \caption{Accuracy of MLLMs at difficulty levels.}
    \label{fig:subfig1}
  \end{subfigure}
  \hfill
  \begin{subfigure}[b]{0.48\textwidth}
    \centering
    \includegraphics[width=0.9\textwidth]{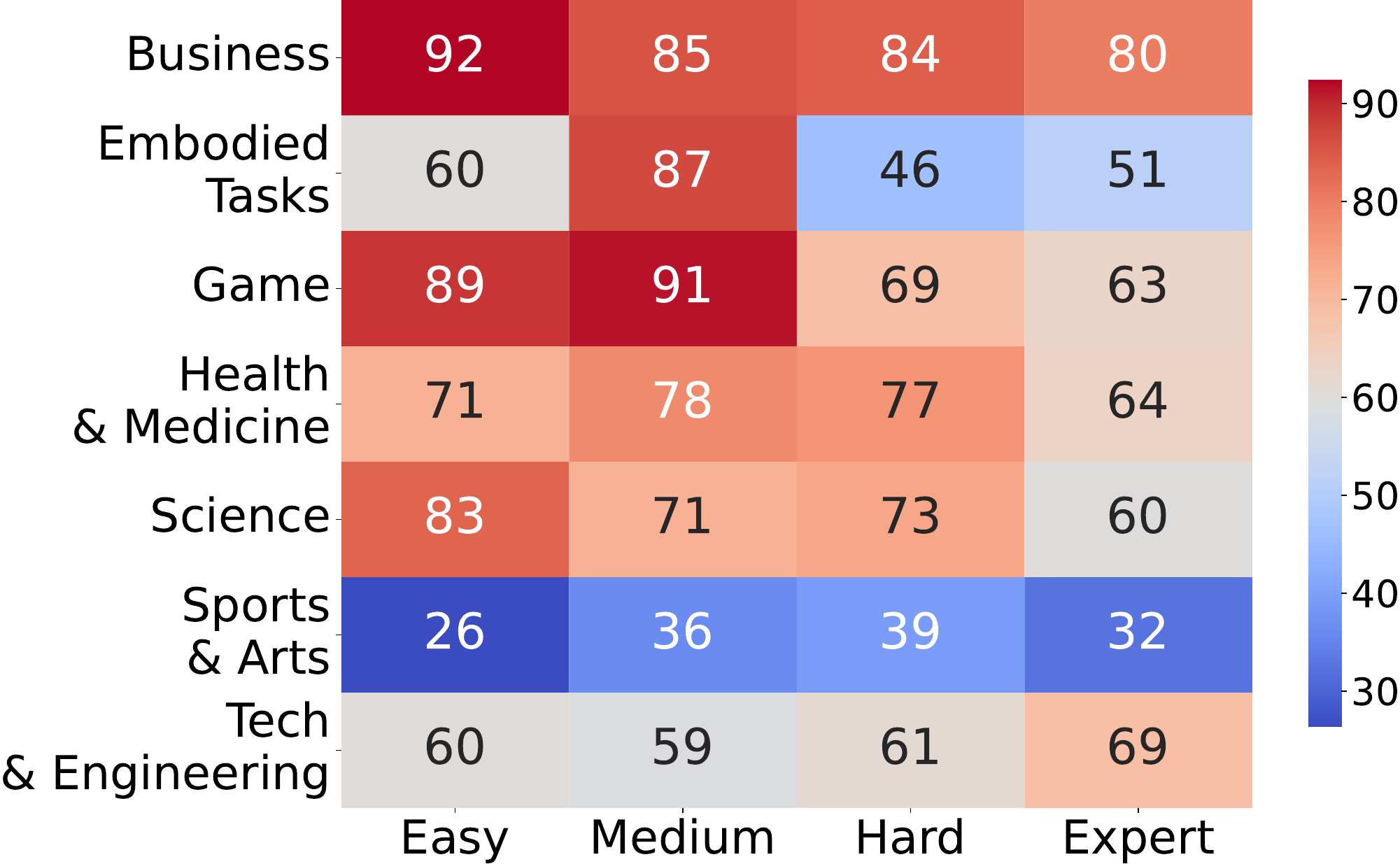}
    \caption{GPT-4V results by disciplines at difficulty levels.}
    \label{fig:subfig2}
  \end{subfigure}
  \caption{Model performance at different difficulty levels for average humans. Average human difficulty levels are defined by 3 turkers' performance per question: Easy (3/3 correct answers), medium (2/3 correct), hard (1/3 correct), and expert (0/3 correct).}
  \label{fig:correlation}
\end{figure}

Figure~\ref{fig:subfig1} indicate some correlation between the difficulty levels as perceived by humans and the performance of MLLMs. MLLMs generally follow a trend where accuracy decreases as the difficulty level increases, which aligns with human performance patterns. However, the correlation is not perfect, suggesting that while models and humans share some common ground in understanding question difficulty, there are also notable differences in their capabilities. The data reveals that MLLMs exhibit different skill sets compared to humans. As highlighted in Figure~\ref{fig:subfig2}, models like GPT-4V can correctly answer expert-level questions that humans often get wrong, particularly in disciplines such as Business and Health \& Medicine, where humans often struggle, yet they sometimes falter on easier questions, likely due to the lack of contextual understanding. Notably, discrepancies in disciplines like Art \& Sports and Tech \& Engineering highlight areas where MLLMs’ performance does not align with human results, suggesting different perception, cognition, and reasoning abilities in handling abstract concepts. These differences suggest that MLLMs can complement human capabilities, offering potential for enhanced task performance by combining the data-driven insights of models with human intuition and contextual knowledge.

\subsection{Study on Modality of Perception}
We conduct ablations to evaluate MLLMs ability to perceiving the world on the synthetic dataset of  \benchmarkname. With our synthetic dataset, we considered scenarios where only one modality—either audio or visual—is available. Table~\ref{audio_only} shows the results which evaluates the model's ability to interpret spoken language, background noises, and other audio elements without the aid of visual context and the model's perception ability to operate without any audio input. For the visual perception test, Gemini Pro performed the best, demonstrating its strong ability to process visual information. Interestingly, Video-Chat exhibited better audio perception than ChatUnivi, despite its poorer visual perception. This may be attributed to its use of the Whisper~\citep{whisper} speech recognition model. It also explains that in Table~\ref{maineval}, Video-Chat outperforms ChatUnivi in the Art \& Sports discipline, which requires a greater understanding of music, voice, and background audio. However, in other disciplines such as Science and Health \& Medicine, Video-Chat's performance is significantly poorer.

\begin{table}[t]
\centering
\caption{Performance on Synthetic Subset I (Audio) and II (Visual). Synthetic Subset I contains QAs based solely on the audio content, while Synthetic Subset II focuses exclusively on the visual content of the video. We evaluated four MLLMs processing both audio and visual inputs along with Gemini Pro (for the audio setting, only providing the question).}
\label{audio_only}
\resizebox{\textwidth}{!}{
\begin{tabular}{lccccccccccccccccc}
\toprule
\multirow{2}{*}{Model}        
& \multicolumn{2}{c}{Art\&Sports} & \multicolumn{2}{c}{Business} & \multicolumn{2}{c}{Science} & \multicolumn{2}{c}{Health\&Medicine} & \multicolumn{2}{c}{Embodied Tasks} & \multicolumn{2}{c}{Tech\&Engineering} & \multicolumn{2}{c}{Game}& \multicolumn{2}{c}{Average} \\
& Audio & Visual & Audio & Visual & Audio & Visual & Audio & Visual & Audio & Visual & Audio & Visual & Audio & Visual & Audio & Visual\\
\midrule
Random Choice & 31.59 & 30.14 & 31.18 & 26.58 & 36.98 & 32.89 & 38.74 & 32.64 & 32.81 & 31.25 & 27.23 & 32.60 & 32.01 & 30.78 & 32.44 & 30.91 \\   
\hdashline
Video-Chat~\citep{li2023videochat} & \textbf{33.98} & 32.48 & \textbf{46.47} & 41.46 & \textbf{41.86} & 39.15 & \textbf{45.95} & 36.81 & 32.81 & 46.88 & \textbf{37.48} & 35.91 & \textbf{32.98}&46.70&\textbf{38.82} & 39.07 \\
ChatUnivi~\citep{jin2023chatunivi} & 30.03 & 43.22 & 30.19 & 52.85 & 38.75 & 54.59 & 34.76 & 50.69 & 20.14 & 40.63 & 24.17 & 46.41 & 29.98& 45.44 & 31.82 & 48.44 \\
Video-LLaMA~\citep{zhang2023videollama} & 30.15 & 30.23 & 36.18 & 33.17 & 31.33 & 31.34 & 30.90 & 32.78 & \textbf{33.13} & 30.05 & 31.18 & 30.55 & 20.49&27.20&29.08 & 30.47 \\
Otter~\citep{li2023otter} & 14.22 & 16.82 & 16.77 & 14.24 & 16.12 & 17.00 & 19.82 & 13.19 & 10.94 & 12.50 & 15.63 & 12.43 & 6.65 &10.44&12.83 & 13.41 \\
Gemini Pro~\citep{team2023gemini} & 20.88 & \textbf{61.38} & 29.43 & \textbf{77.35} & 30.62 & \textbf{74.26} & 30.14 & \textbf{81.53} & 22.57 & \textbf{70.31} & 18.83 & \textbf{66.22} & 29.96 & \textbf{65.01} & 24.45 &\textbf{69.97} \\
\bottomrule
\end{tabular}}
\end{table}

\subsection{Error Analysis}
\begin{figure}[!t]
    \centering
    \includegraphics[width=\textwidth]{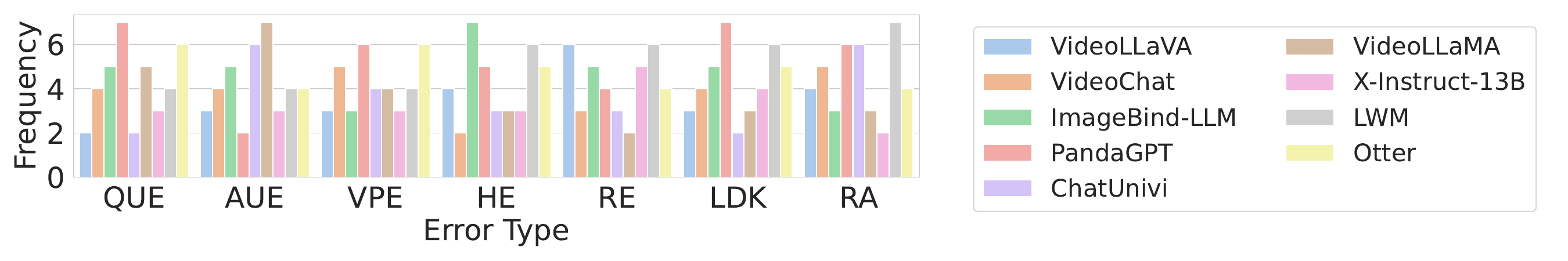}
    \caption{ The frequency of different error types across various MLLMs. For each error type, 10 examples were evaluated. Error types are abbreviated as follows: QUE (Question Understanding Error), AUE (Audio Understanding Error), VPE (Visual Perception Error), HE (Hallucination Error), RE (Reasoning Error), LDK (Lack of Domain Knowledge), and RA (Reject to Answer).}
    \label{fig:error_analysis}
\end{figure}

To gain deeper insights into the limitations of MLLMs, we prompted the models to explain the reasoning behind their choices, particularly when errors occurred. Through this analysis, we identified common error patterns and summarized them into seven distinct categories. We conducted a simple test where the same questions that triggered errors in GPT-4V were also posed to other MLLMs. The frequencies of each type of error are presented in Figure~\ref{fig:error_analysis}, as annotated by human evaluators. Detailed qualitative examples of these errors and further analysis are provided in the Appendix.

\section{Conclusion}
\label{sec:conclusion}
Our~\benchmarkname Benchmark represents a significant step forward in the quest for advanced multi-modal language models capable of understanding complex video content. By presenting a diverse array of videos across seven disciplines, accompanied by questions that challenge models to demonstrate explanation, counterfactual thinking, future prediction, and domain expertise, we have created a rigorous testing ground for the next generation of AI. While using LLMs for data generation can introduce hallucination issues, these challenges are manageable and are commonly addressed~\citep{do_not_answer,Do_Anything_Now}. Another potential risk is the misuse of MLLMs for surveillance or privacy invasion. The ability of models to understand video content and perform reasoning could be exploited to monitor individuals without their consent, leading to serious ethical and legal concerns regarding privacy.

\bibliographystyle{plainnat}
\bibliography{main}

\appendix

\section{Overview of the Appendix}

We host the project website on \url{https://mmworld-bench.github.io/}. The benchmark and code implementations can be found at \url{https://github.com/eric-ai-lab/MMWorld}. The link to Croissant metadata record documenting the dataset/benchmark available for viewing and downloading is available at \url{https://github.com/eric-ai-lab/MMWorld/blob/main/data/croissanta_hf_data.json}.
This Appendix is organized as follows:
\begin{itemize}
    \item Section~\ref{results} contains additional experimental results;
    \item Section~\ref{impelmentation} contains the implementation details;
    \item Section~\ref{appendix:human_eval} contains the settings and results from human evaluations;
    \item Section~\ref{error} contains the error analysis;
    \item Section~\ref{data} contains the data examples from \benchmarkname;
    \item Section~\ref{subdiscipline_statistics} contains additional data statistics of \benchmarkname;
     \item Section~\ref{sec:datasheet} contains the datasheet of \benchmarkname;
      \item Section~\ref{sec:license_host} contains the author statement, licence, and maintenance plan.
\end{itemize}

\section{Additional Results}
\label{results}

\subsection{Results Across Different Seed for Each Model}
In Table~\ref{maineval_three_seeds}, we show detailed results using three different seeds for each evaluated models.
\begin{table}[th]
\centering
\caption{Detailed results of model performance, measured as accuracy percentages across diverse disciplines for three runs. The random choice baseline involves shuffling candidate answers for each video question before consistently selecting answer `a'. GPT-4V and Gemini Pro utilize 10 image frames extracted from the video content. }
\label{maineval_three_seeds}
\resizebox{\linewidth}{!}{
\begin{tabular}{lcccccccc}
\toprule
\multirow{2}{*}{Model}        
& {Art\& } & \multirow{2}{*}{Business} & \multirow{2}{*}{Science} & {Health\&} & {Embodied } & {Tech\& } & \multirow{2}{*}{Game }& \multirow{2}{*}{Average} \\
& Sports& & &  Medicine& Tasks& Engineering& \\ \hline 
GPT-4o-seed 1~\citep{gpt4o} & 47.10 & 92.31 & 75.11 & 81.25 & 65.03 & 72.34 & 78.31 & 62.22 \\
GPT-4o-seed 2~\citep{gpt4o} & 46.58 & 90.91 & 69.78 & 84.38 & 65.73 & 75.53 & 83.13 & 61.77 \\
GPT-4o-seed 3~\citep{gpt4o} & 49.94 & 90.21 & 76.44 & 84.38 & 58.04 & 78.72 & 79.52 & 63.63 \\
Claude-3.5-seed 1~\citep{claude} & 54.32           & 64.34           & 59.11           & 53.12           & 30.77           & 59.57           & 59.04           & 54.27 \\
Claude-3.5-seed 2~\citep{claude} & 54.32           & 63.64           & 61.33           & 54.17           & 30.77           & 58.51           & 59.04           & 54.52 \\
Claude-3.5-seed 3~\citep{claude} & 55.10           & 63.64           & 59.11           & 56.25           & 31.47           & 58.51           & 60.24           & 54.84 \\
GPT-4V-seed 1~\citep{gpt4-v} & 36.90 & 79.72 & 64.00 & 73.96 & 51.75 & 60.64 & 71.08 & 51.64 \\
GPT-4V-seed 2~\citep{gpt4-v} & 35.48 & 83.92 & 68.44 & 73.96 & 58.04 & 60.64 & 75.90 & 52.79 \\
GPT-4V-seed 3~\citep{gpt4-v} & 36.13 & 81.12 & 67.11 & 72.92 & 56.64 & 62.77 & 73.49 & 52.47 \\
Gemini Pro-seed 1~\citep{team2023gemini} & 40.90 & 79.72 & 60.44 & 78.12 & 43.36 & 71.28 & 65.06 & 52.92 \\
Gemini Pro-seed 2~\citep{team2023gemini} & 35.10 & 75.52 & 63.11 & 75.00 & 44.06 & 71.28 & 69.88 & 50.16 \\
Gemini Pro-seed 3~\citep{team2023gemini} & 35.35 & 74.83 & 64.89 & 77.08 & 43.36 & 67.02 & 63.86 & 49.97 \\
Video-LLaVA-seed 1~\citep{videollava} & 34.58 & 51.05 & 57.33 & 32.29 & 61.54 & 57.45 & 50.60 & 43.94 \\
Video-LLaVA-seed 2~\citep{videollava} & 36.77 & 52.45 & 56.00 & 32.29 & 65.03 & 57.45 & 51.81 & 45.35 \\
Video-LLaVA-seed 3~\citep{videollava} & 36.39 & 50.35 & 55.56 & 33.33 & 62.94 & 59.57 & 44.58 & 44.52 \\
Video-Chat-seed 1~\citep{li2023videochat} & 39.48 & 51.05 & 30.67 & 46.88 & 39.86 & 39.36 & 44.58 & 40.03 \\
Video-Chat-seed 2~\citep{li2023videochat} & 39.48 & 51.05 & 30.67 & 45.83 & 41.26 & 39.36 & 45.78 & 40.15 \\
Video-Chat-seed 3~\citep{li2023videochat} & 39.61 & 51.05 & 31.11 & 45.83 & 40.56 & 39.36 & 44.58 & 40.15 \\
mPLUG-Owl-seed 1 ~\citep{ye2023mplug} & 31.35 & 65.73 & 45.78 & 61.46 & 28.67 & 48.94 & 65.06 & 41.05 \\
mPLUG-Owl-seed 2~\citep{ye2023mplug} & 28.65 & 65.03 & 44.44 & 58.33 & 21.68 & 37.23 & 57.83 & 37.52 \\
mPLUG-Owl-seed 3 ~\citep{ye2023mplug} & 27.48 & 61.54 & 52.00 & 60.42 & 20.98 & 39.36 & 63.86 & 38.23 \\
ChatUnivi-seed 1~\citep{jin2023chatunivi} & 24.13 & 60.14 & 52.00 & 62.50 & 48.95 & 56.38 & 56.63 & 39.77 \\
ChatUnivi-seed 2~\citep{jin2023chatunivi} & 25.16 & 62.94 & 51.11 & 62.50 & 44.06 & 58.51 & 50.60 & 39.77 \\
ChatUnivi-seed 3~\citep{jin2023chatunivi} & 24.13 & 59.44 & 52.89 & 58.33 & 45.45 & 55.32 & 50.60 & 38.87 \\
PandaGPT-seed 1~\citep{su2023pandagpt} & 26.06 & 44.06 & 38.22 & 41.67 & 35.66 & 39.36 & 42.17 & 32.97 \\
PandaGPT-seed 2~\citep{su2023pandagpt} & 24.77 & 45.45 & 36.89 & 34.38 & 34.27 & 40.43 & 44.58 & 31.88 \\
PandaGPT-seed 3~\citep{su2023pandagpt} & 25.16 & 38.46 & 43.11 & 39.58 & 36.36 & 45.74 & 33.73 & 32.58 \\
ImageBind-LLM-seed 1~\citep{han2023imagebind} & 24.77 & 41.96 & 30.67 & 31.25 & 46.85 & 43.62 & 40.96 & 31.62 \\
ImageBind-LLM-seed 2~\citep{han2023imagebind} & 25.03 & 41.96 & 32.44 & 31.25 & 45.45 & 40.43 & 40.96 & 31.69 \\
ImageBind-LLM-seed 3~\citep{han2023imagebind}& 24.65 & 44.06 & 33.33 & 28.12 & 48.25 & 40.43 & 42.17 & 31.94 \\
X-Instruct-BLIP-seed 1~\citep{panagopoulou2023xinstructblip} & 21.42 & 14.69 & 22.22 & 29.17 & 16.78 & 21.28 & 26.51 & 21.23 \\
X-Instruct-BLIP-seed 2~\citep{panagopoulou2023xinstructblip} & 20.77 & 16.78 & 24.00 & 28.12 & 20.28 & 22.34 & 25.30 & 21.62 \\
X-Instruct-BLIP-seed 3~\citep{panagopoulou2023xinstructblip} & 21.03 & 16.08 & 21.33 & 28.12 & 18.18 & 23.40 & 26.51 & 21.23 \\
LWM-seed 1~\citep{lwm} & 11.35 & 18.18 & 16.44 & 19.79 & 24.48 & 24.47 & 10.84 & 15.20 \\
LWM-seed 2~\citep{lwm} & 12.13 & 17.48 & 15.56 & 19.79 & 24.48 & 22.34 & 8.43 & 15.14 \\
LWM-seed 3~\citep{lwm} & 12.65 & 16.78 & 14.22 & 21.88 & 28.67 & 19.15 & 15.66 & 15.84 \\
Otter-seed 1~\citep{li2023otter} & 18.45 & 19.58 & 8.89 & 8.33 & 14.69 & 15.96 & 14.46 & 15.84 \\
Otter-seed 2~\citep{li2023otter} & 17.29 & 17.48 & 9.33 & 6.25 & 13.99 & 18.09 & 15.66 & 15.14 \\
Otter-seed 3~\citep{li2023otter}  & 15.61 & 18.88 & 9.78 & 6.25 & 11.19 & 13.83 & 15.66 & 13.98 \\
Video-LLaMA-seed 1~\citep{zhang2023videollama}  & 5.55 & 21.68 & 24.00 & 29.17 & 15.38 & 21.28 & 18.07 & 13.66 \\
Video-LLaMA-seed 2~\citep{zhang2023videollama}  & 6.58 & 20.28 & 20.44 & 31.25 & 13.99 & 17.02 & 32.53 & 14.05 \\
Video-LLaMA-seed 3~\citep{zhang2023videollama}  & 6.32 & 21.68 & 22.22 & 33.33 & 16.78 & 19.15 & 24.10 & 14.37 \\
 \bottomrule
\end{tabular}}
\end{table}

\subsection{Results from Amazon Turkers}
Table~\ref{turker_results} presents the evaluation results from three sets of Amazon Turkers across various disciplines. The results indicate that there is slightly variability in performance across different human evaluators.
\begin{table}[t]
\centering
\caption{Performance of different set of turkers}
\label{turker_results}
\resizebox{\linewidth}{!}{
\begin{tabular}{lcccccccc}
\toprule
\multirow{2}{*}{Model}        
& {Art\& } & \multirow{2}{*}{Business} & \multirow{2}{*}{Science} & {Health\&} & {Embodied } & {Tech\& } & \multirow{2}{*}{Game\& }& \multirow{2}{*}{Average} \\
& Sports& & &  Medicine& Tasks& Engineering&  \\ \hline
Turker Set 1 & 25.224 & 39.860 & 32.444 & 40.625 & 51.049 & 50.000 & 40.964 & 33.227 \\ 
Turker Set 2 &30.452 & 46.154 & 35.556 & 42.708 & 53.846 & 51.064 & 46.988 & 37.652 \\
Turker Set 3 &26.710 & 41.958 & 36.889 & 46.875 & 53.147 & 42.553 & 38.554 & 34.830 \\
\bottomrule
\end{tabular}}
\end{table}

\subsection{Results for the Two Different Evaluation Strategies}
\begin{table}[t]
\centering
\caption{Performance of different MLLMs across different disciplines.}
\label{maineval_two_strategy}
\resizebox{\linewidth}{!}{
\begin{tabular}{lccccccc}
\toprule
\multirow{2}{*}{Model}        
& {Art\& } & \multirow{2}{*}{Business} & \multirow{2}{*}{Science} & {Health\&} & {Embodied } & {Tech\& } & \multirow{2}{*}{Average} \\
& Sports& & &  Medicine& Tasks& Engineering&  \\ \hline
Video-Chat (Open-ended)~\citep{li2023videochat}&27.484           & 9.091           & 18.137           & 10.417           & 29.371           & 19.149           & 22.887 \\
Video-Chat ~\citep{li2023videochat}& 39.355           & 48.951           & 31.863           & 45.833           & 39.161           & 38.298           & 39.588\\
Video-LLaMA (Open-ended)~\citep{zhang2023videollama}      & 5.419           & 27.972           & 24.020           & 31.250           & 11.816          & 15.957           & 16.096\\
Video-LLaMA~\citep{zhang2023videollama}      & 27.355           & 31.469           & 31.373           & 48.958           & 16.084           & 28.723           & 28.729              \\
ChatUnivi (Open-ended) ~\citep{jin2023chatunivi}       & 21.161           & 61.538           & 42.157           & 61.458           & 30.070           & 37.234           & 32.646               \\
ChatUnivi ~\citep{jin2023chatunivi}       & 12.387           & 58.042           & 50.000           & 60.417           & 30.070           & 43.617           & 29.072 \\
Otter (Open-ended)~\citep{li2023otter}  &37.677           & 32.867           & 37.255           & 32.292           & 22.378           & 27.660           & 34.639 \\
Otter~\citep{li2023otter}  &17.677           & 16.783           & 12.255           & 5.208           & 17.483           & 15.957           & 15.876  \\
ImageBind-LLM (Open-ended)~\citep{han2023imagebind}    & 3.355           & 3.497           & 14.706           & 10.417           & 21.678           & 18.085           & 8.179    \\
ImageBind-LLM~\citep{han2023imagebind}  &23.742           & 34.965           & 51.471           & 33.333           & 48.951           & 56.383           & 33.952 \\
PandaGPT (Open-ended)~\citep{su2023pandagpt}       & 22.581           & 16.084           & 24.020           & 21.875           & 19.580           & 21.277           & 21.718                \\
PandaGPT~\citep{su2023pandagpt}  & 27.613           & 44.056           & 39.706           & 25.000           & 40.559           & 21.277           & 31.615\\
LWM (Open-ended)~\citep{lwm} & 16.000           & 20.979           & 14.706           & 16.667           & 19.580           & 20.213           & 16.976       \\
LWM~\citep{lwm}              & 16.387           & 18.182           & 18.137           & 19.792           & 22.378           & 21.277           & 17.938         \\
X-Instruct-BLIP (Open-ended)~\citep{panagopoulou2023xinstructblip}   & 3.613           & 11.888           & 14.706           & 25.000           & 17.483           & 13.830           & 9.416                \\
X-Instruct-BLIP~\citep{panagopoulou2023xinstructblip} &19.355           & 13.287           & 22.549           & 29.167           & 18.881           & 14.894           & 19.519\\
 \bottomrule
\end{tabular}}
\end{table}

In Table~\ref{maineval_two_strategy}, we give additional evaluation results for different MLLMs evaluated in this paper. For closed-source models, the evaluation pipeline is the one used in the main paper, which involves utilizing GPT-4V as a judger. The process consists of presenting GPT-4V with the question, a corresponding answer generated by the baseline model, and the set of possible options. GPT-4V then assesses whether the model-generated answer is accurate within the given context; Another is open-ended generation where we employ a two-step methodology. We first prompt each model to do open-ended generation. Subsequently, we prompt the model to align its generative response with one of the predefined options: `a', `b', `c', or `d'. 

\subsection{Detailed Results on Multi-faceted Reasoning}

\begin{table}[t]
\centering
\caption{Detailed results of different MLLMs on multi-faceted reasoning.}
\label{multifacetedreasoning}
\resizebox{\linewidth}{!}{
\begin{tabular}{lcccccc}
\toprule
\multirow{2}{*}{Model} & \multirow{2}{*}{Explanation} & Counterfactual & Future & Domain & Attribution & Temporal \\
 &  & Thinking & Prediction & Expertise & Understanding & Understanding \\
\midrule
\multicolumn{7}{c}{\textit{Proprietary Models}} \\
\midrule
GPT-4V & 44.90 & 64.90 & \textbf{78.59} & \textbf{61.07} & \textbf{59.61} & 27.17 \\
Gemini Pro & \textbf{48.58} & \textbf{65.49} & 65.45 & 53.87 & 43.92 & 24.65 \\
\midrule 
\multicolumn{7}{c}{\textit{Open-source Models}} \\
\midrule
Video-LLaVA-7B & 42.46 & 42.55 & 64.96 & 47.86 & 36.86 & \textbf{34.45} \\
VideoChat-7B & 41.66 & 43.73 & 45.74 & 40.95 & 30.59 & 25.77 \\
ImageBind-LLM-7B & 29.51 & 26.86 & 50.61 & 33.93 & 34.90 & 19.89 \\
PandaGPT-7B & 29.55 & 37.45 & 46.47 & 33.93 & 26.27 & 28.01 \\
ChatUnivi-7B & 33.91 & 48.82 & 61.80 & 45.95 & 33.33 & 22.97 \\
VideoLLaMA-2-13B & 10.55 & 23.92 & 25.30 & 16.31 & 8.63 & 6.16 \\
X-Instruct-BLIP-7B & 23.05 & 15.29 & 27.25 & 21.07 & 24.31 & 11.20 \\
LWM-1M-JAX & 11.62 & 18.82 & 30.66 & 17.98 & 21.57 & 7.00 \\
Otter-7B & 16.91 & 10.98 & 15.82 & 13.10 & 17.65 & 9.52 \\
mPLUG-Owl-7B & 35.20 & 49.61 & 55.47 & 47.74 & 24.71 & 20.17 \\
\bottomrule
\end{tabular}}
\end{table}

In Table~\ref{multifacetedreasoning}, we give detailed performance numbers of different MLLMs on multi-faceted reasoning corresponding to Figure 4 in the main paper.

\section{Implementation Details}
\label{impelmentation}
We use the optimum number of video frames and report the performance in the main paper. The numbers of the sampled frames are 10 for GPT-4V/o and Gemini Pro, 8 for Video-LLaVA, 32 for ChatUniVi. For closed-source models, for both Gemini Pro and GPT-4V, we use the default settings provided by their official APIs. We use Katna~\footnote{
https://github.com/keplerlab/katna} to extract key video frames as input to these two models. The Gemini Pro is set to process visual input and configured with safety settings to filter a range of harmful content. The configuration thresholds are set to `BLOCK\_NONE'. For PandaGPT, we set `top\_p' to 0.7, and `temperature' to 0.5. For VideoChat, we set `max\_frames' to 100. For LWM, we use the LWM-Chat-1M variant. For X-Instruct-BLIP, the model is implemented using four image frames. For Otter, we use the video variant. We use GPT-4-32K as the judge for judging whether the model answer is correct when it can not mapped to the option letter using the rule-based method. The prompt provided to GPT-4-32K is structured as follows: \texttt{"I will present a response from a question-answering model alongside several answer options. Your task is to evaluate the response and determine which of the following options it most closely aligns with, denoting the most similar option by its corresponding letter (a, b, c, or d)."}.

\paragraph{Query Generation in Synthetic Data Generation Pipeline}
For the discipline of \textbf{Science}, queries are generated for subdisciplines such as Geography, Chemistry, Wildlife Restoration, Mycology, Nature, Physics, Weather, Zoology, Math, Botany, Biology, and Geology. In the \textbf{Tech \& Engineering} discipline, our queries span across Electronics, Animal Behavior, Mechanical Engineering, Energy \& Power, Architecture, Agriculture, Nature, Physics, Robotics, Woodworking, and Gardening. The \textbf{Sports \& Arts} discipline encompasses a broad range of cultural and physical activities, including Music, Drawing and Painting, Football, Volleyball, Aerobic Gymnastics, Basketball, Instrument, Baking, Dance, Woodworking, Graffiti, Anatomy, and additional Music-related topics. \textbf{Embodied Tasks} are represented through queries for Assembly, Ego-motion, and Single Object Manipulation, focusing on the interaction between agents and their physical environment. The \textbf{Health \& Medicine} discipline is segmented into Pharmacy, Public Health, Clinical Medicine, and Basic Medical Science, reflecting the multifaceted nature of healthcare and medical studies.
The~\textbf{Business} discipline is stratified into fundamental areas such as accounting, finance, management, marketing, and economics, each representing key facets of the commercial and economic world. Lastly, the \textbf{Game} discipline consists of Role Playing Game, First Person Shooting game, Racing Game, Adventure Game, Real-Time Strategy Game, Tower Defense game, and Fighting Game.

Each generated query retrieves relevant video content, which is then filtered and processed to align with the specific needs of our research objectives. Videos that meet our criteria in terms of content, length, and quality are downloaded and incorporated into our dataset, forming the basis for subsequent analysis and model training.

\begin{figure}[tb]
  \centering
  \includegraphics[width=0.8\textwidth]{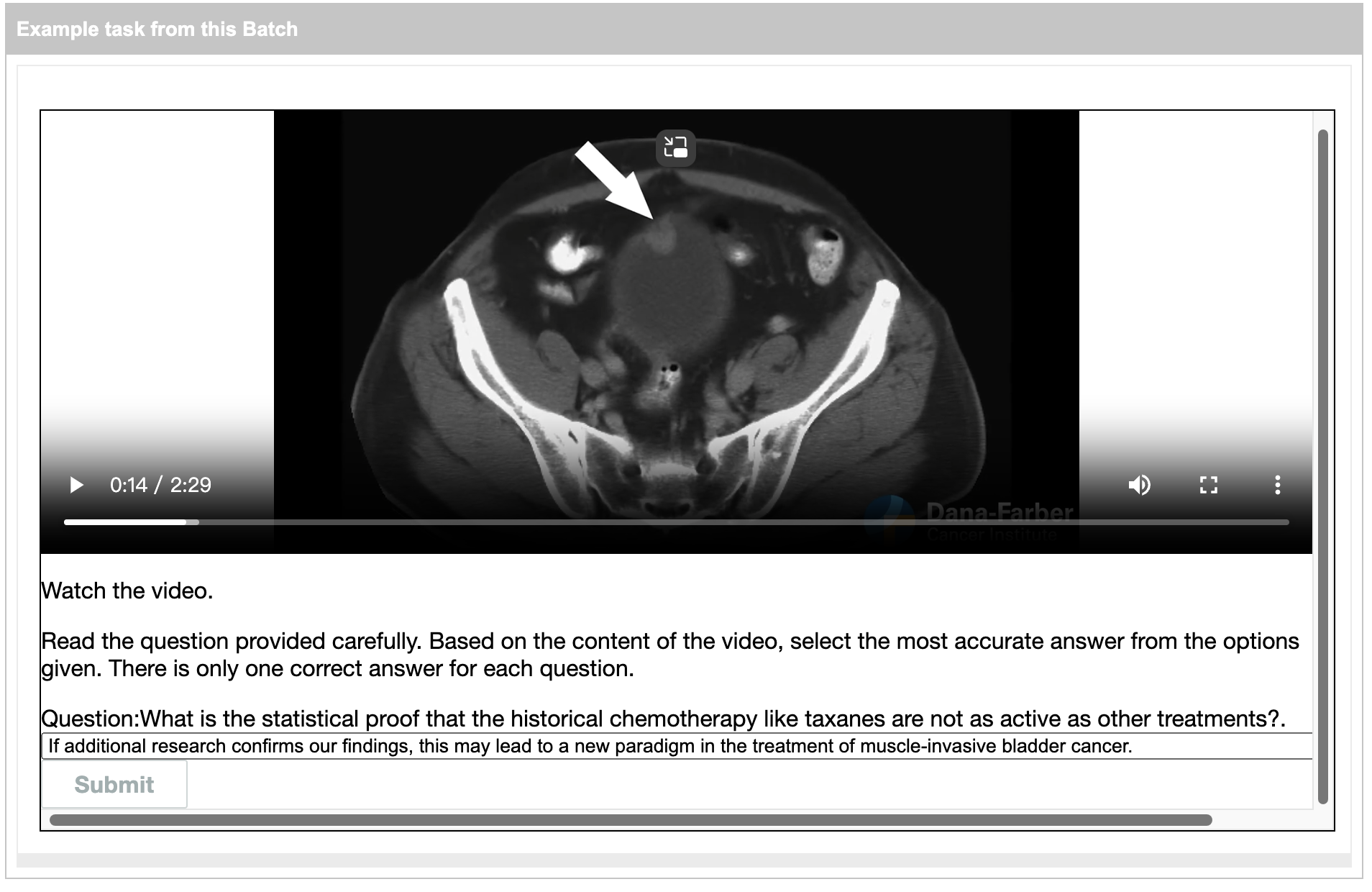}
  \caption{The interface of using Amazon
Mechanical Turk to do human evaluation.
  }
  \label{amt_img}
\end{figure}

\begin{figure}[tb]
  \centering
  \includegraphics[width=0.8\textwidth]{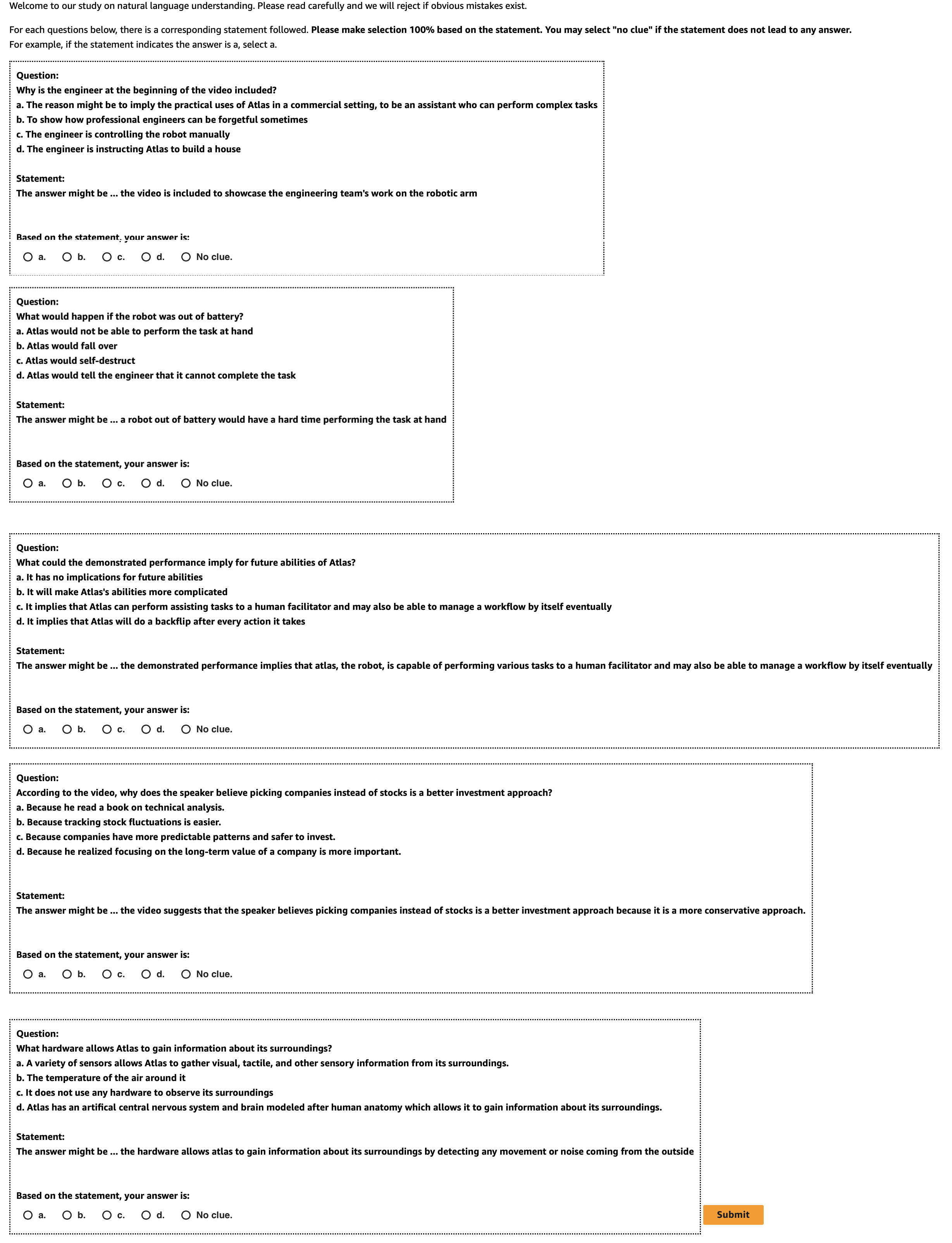}
  \caption{Human evaluation interface for GPT judger.
  }
  \label{fig:human_eval}
\end{figure}

\begin{table}[t]
\centering
\caption{Category-wise and overall error rates}
\begin{tabular}{lcc}
\toprule
\textbf{Category} & \textbf{Incorrect/Total} & \textbf{Error Rate (\%)} \\
\hline
Sports \& Arts & 5/62 & 8.06 \\
Health \& Medicine & 2/7 & 28.57 \\
Science & 1/52 & 1.92 \\
Robotics & 0/12 & 0.00 \\
Business & 0/10 & 0.00 \\
Tech \& Engineering & 1/46 & 2.17 \\
\hline
\textbf{Overall} & \textbf{9/189} & \textbf{4.76} \\
\bottomrule
\end{tabular}
\label{tab:error_rates}
\end{table}

\begin{figure}[t]
  \centering
  \includegraphics[width=\textwidth]{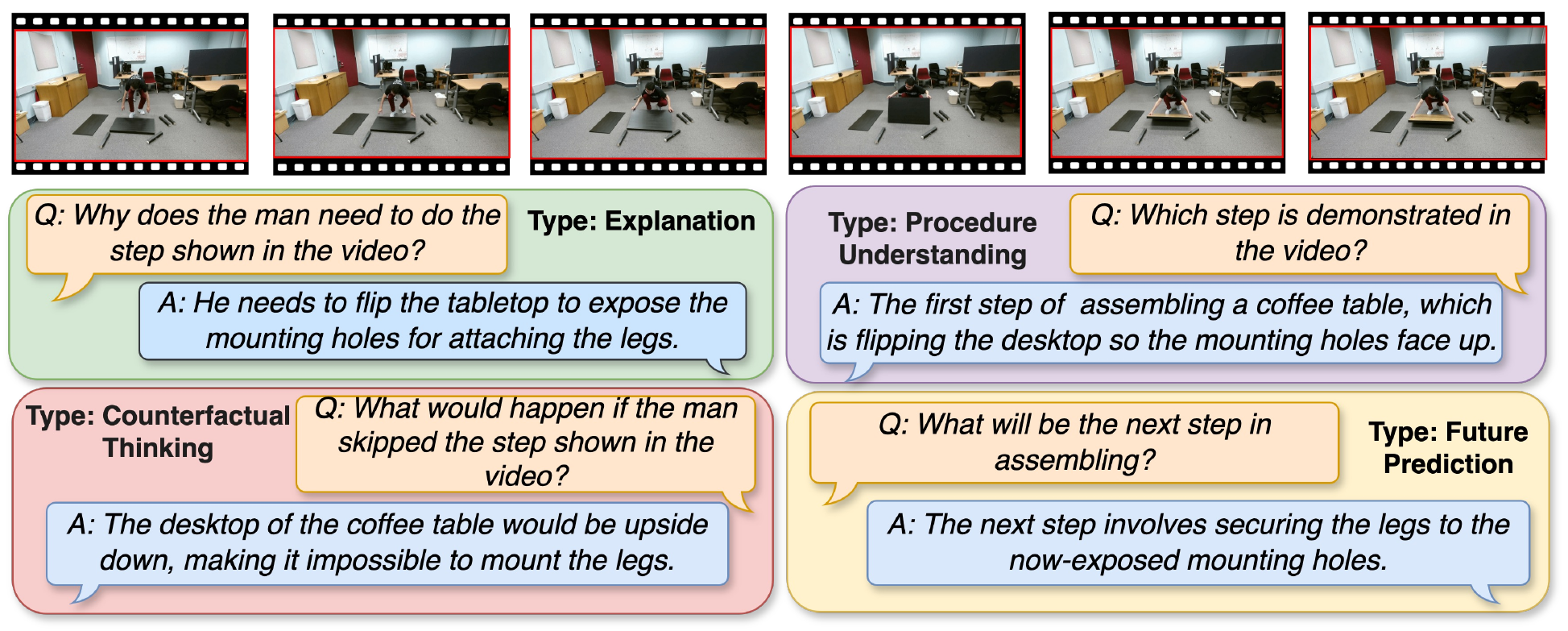}
  \caption{Examples from \benchmarkname in the Embodied Tasks discipline.
  }
  \label{fig:data_examples1}
\end{figure}

\begin{figure}[t]
  \centering
  \includegraphics[width=\textwidth]{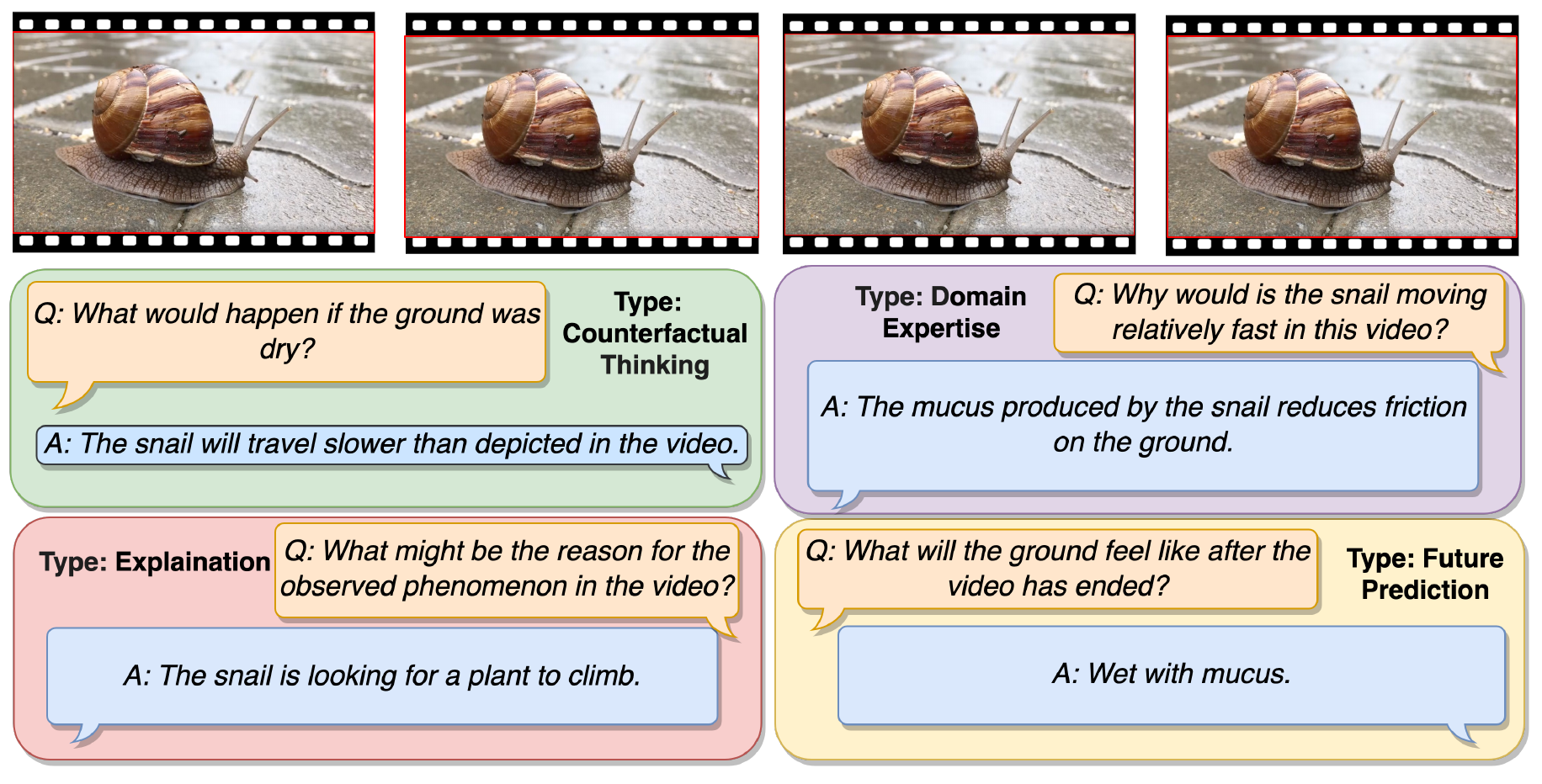}
  \caption{Examples from \benchmarkname in the Tech \& Engineering discipline.
  }
  \label{fig:data_examples2}
\end{figure}

\begin{figure}[t]
  \centering
  \includegraphics[width=\textwidth]{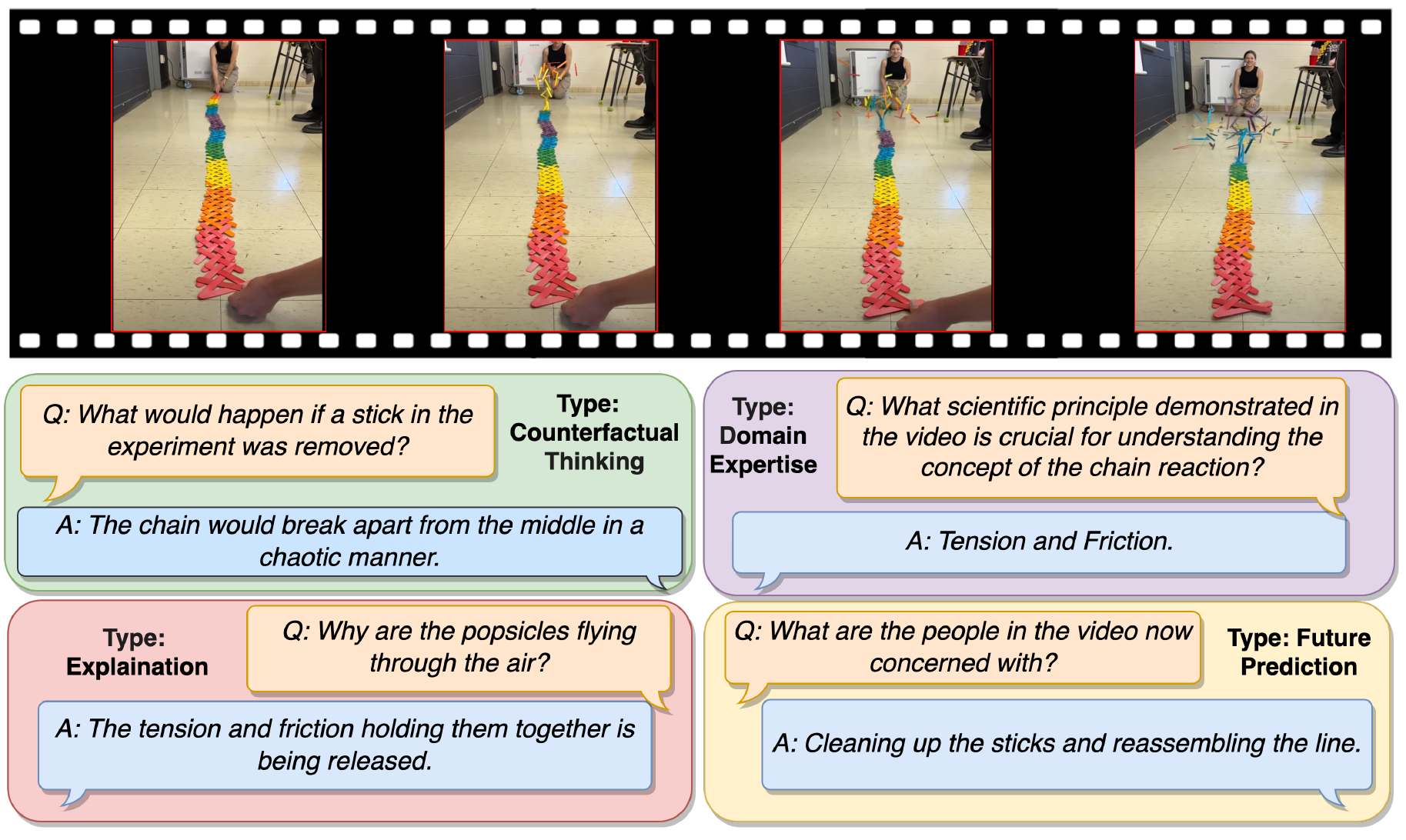}
  \caption{Examples from \benchmarkname in the Science discipline.
  }
  \label{fig:data_examples3}
\end{figure}

\begin{figure}[t]
  \centering
  \includegraphics[width=\textwidth]{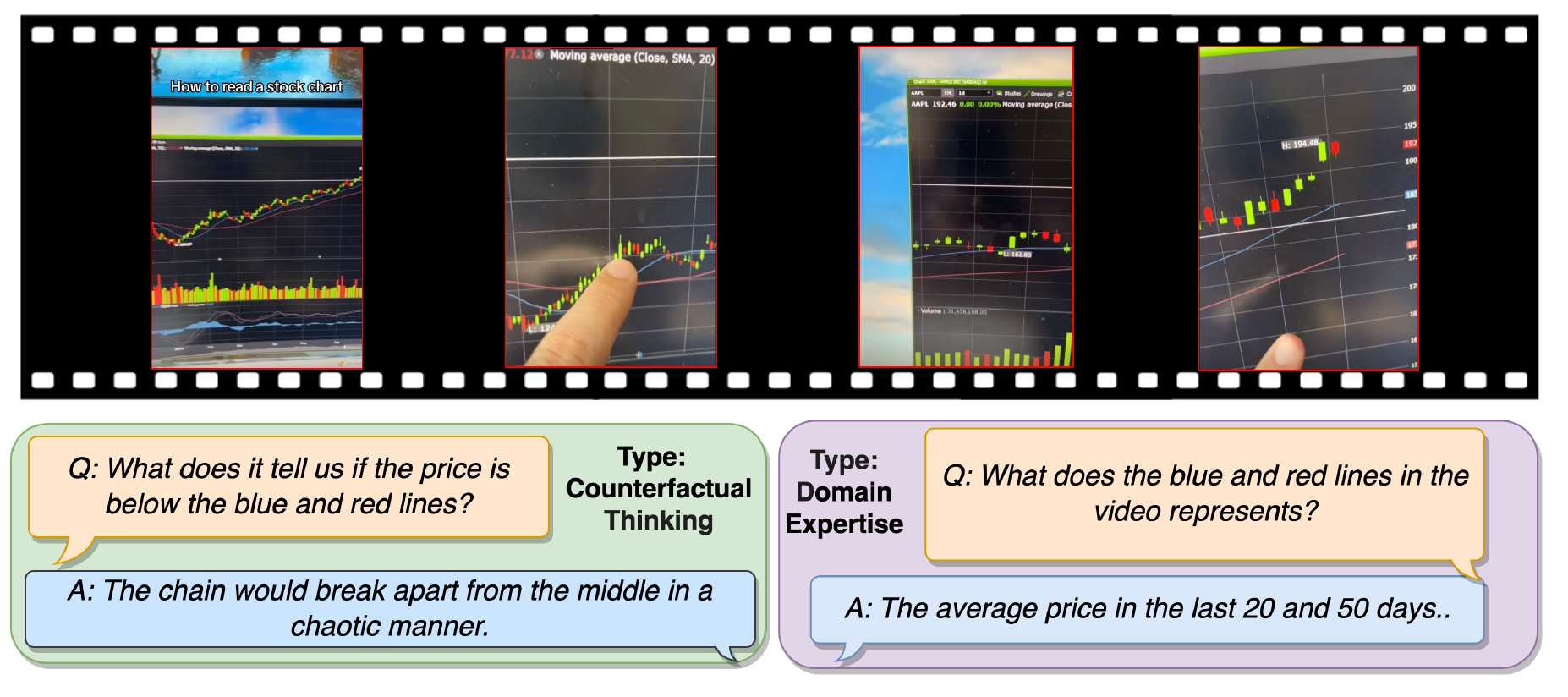}
  \caption{Examples from \benchmarkname in the Business discipline.
  }
  \label{fig:data_examples4}
\end{figure}

\begin{figure}[t]
  \centering
  \includegraphics[width=\textwidth]{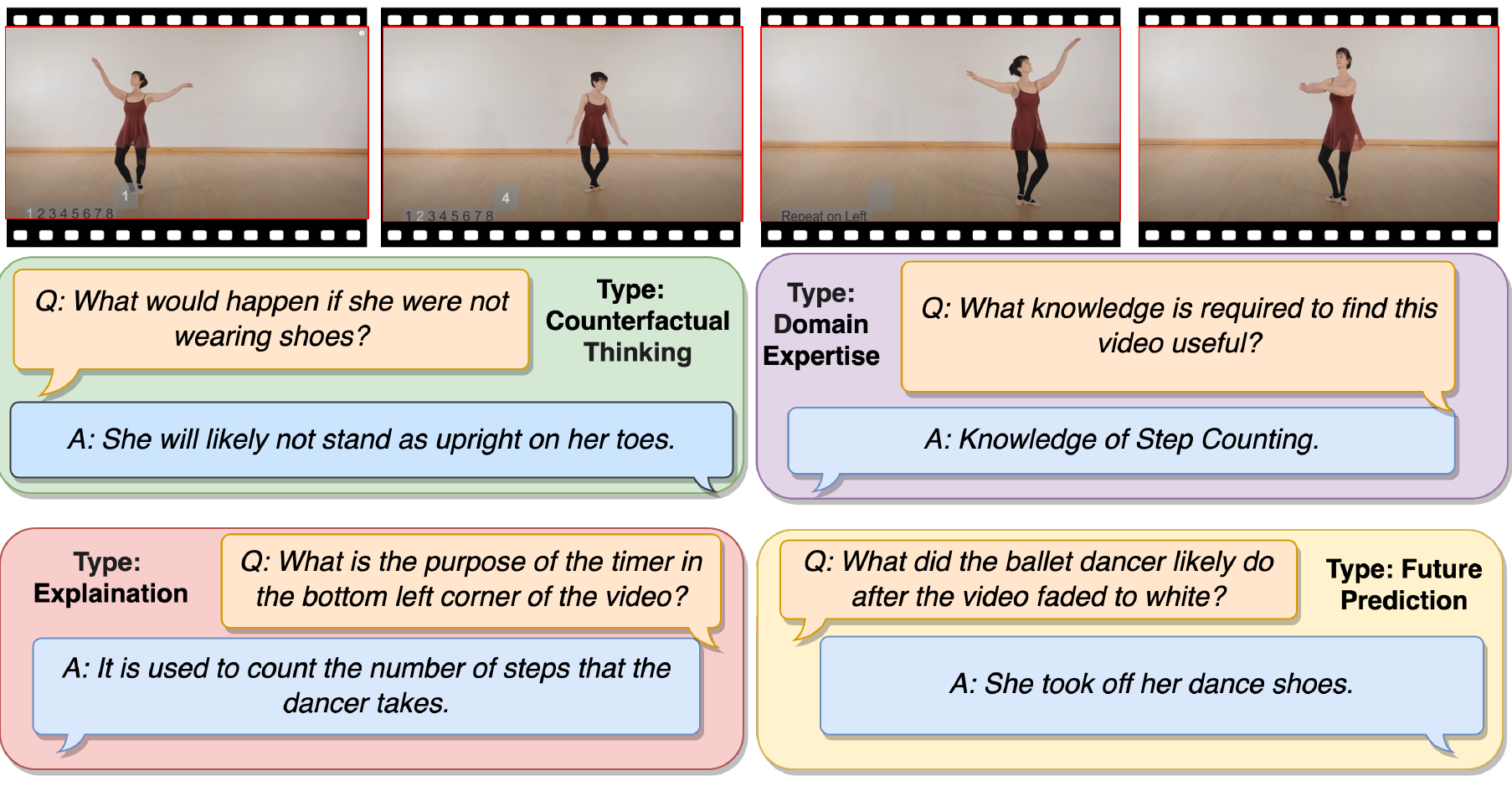}
  \caption{Examples from \benchmarkname in the Arts \& Sports discipline.
  }
  \label{fig:data_examples5}
\end{figure}

\begin{figure}[tb]
  \centering
  \includegraphics[width=\textwidth]{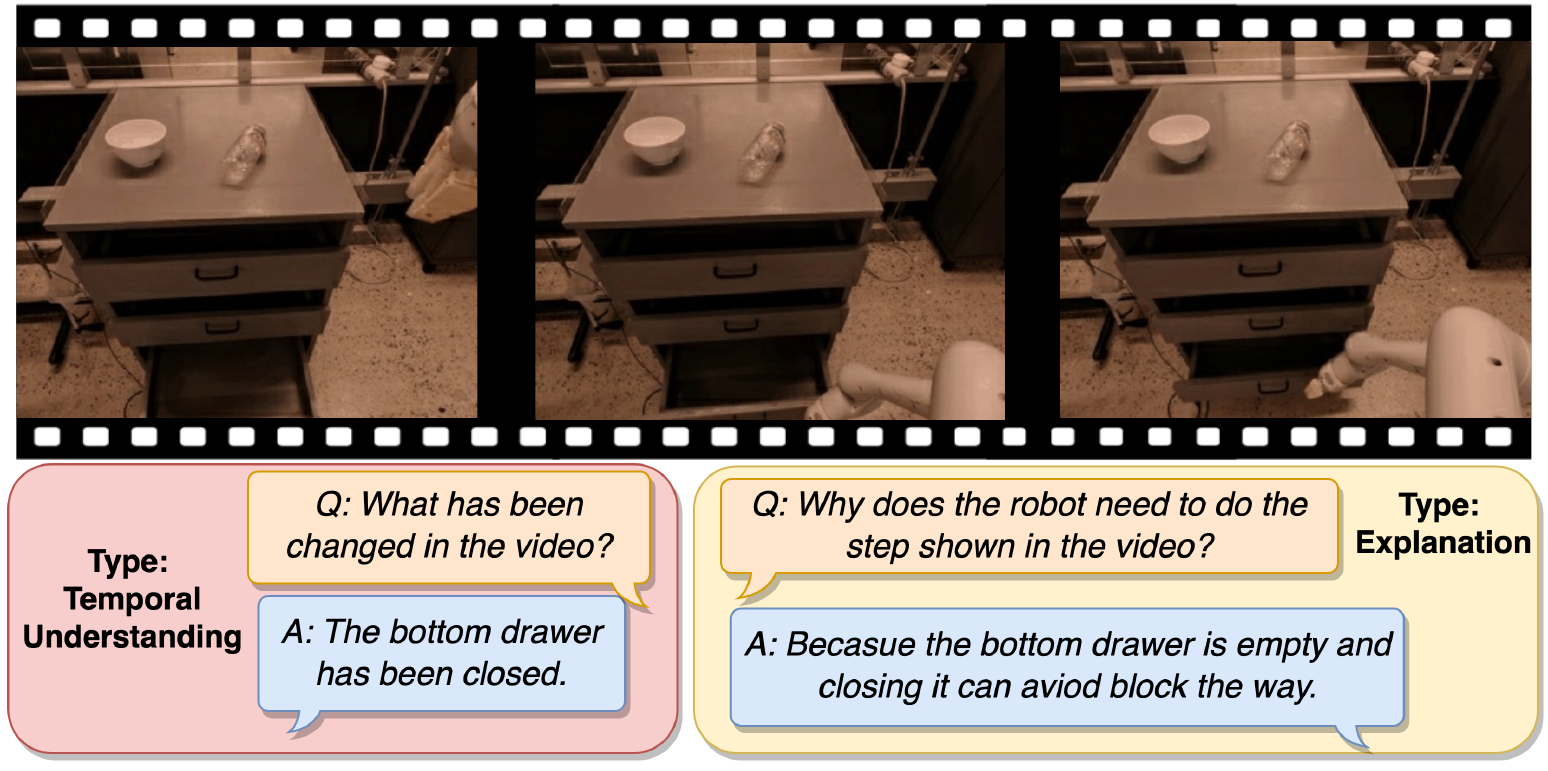}
  \caption{Examples from \benchmarkname of explicit temporal understanding and implicit temporal understanding (e.g., in explanation).
  }
  \label{fig:temp_example}
\end{figure}

\section{Human Evaluation}
\label{appendix:human_eval}
\subsection{Quality of Data}
We hired Amazon Mechanical Turk to do human evaluation on the data with the results shown in Table~\ref{turker_results}. Workers were required to have completed more than 1000 Human Intelligence Tasks (HITs) and have an HIT approval rate greater than 95\% to qualify for our tasks. We show in Figure~\ref{amt_img} the human evaluation interface on the generated data. Each worker was compensated $0.20$ for completing an assignment. This amount was determined based on the estimated time and effort required to complete each task. We set the number of unique workers per task to 3 to collect diverse perspectives while avoiding redundancy. Workers were given 1 hour to complete each assignment. This time frame was chosen to enable thoughtful responses from workers.

\begin{table}[t]
\centering
\caption{Comparison of Human Evaluation on subset of 75 videos.}
\resizebox{\linewidth}{!}{
\begin{tabular}{lccccccc}
\toprule
\multirow{2}{*}{Model}        
& {Art\& } & \multirow{2}{*}{Business} & \multirow{2}{*}{Science} & {Health\&} & {Embodied } & {Tech\& } & \multirow{2}{*}{Average} \\
& Sports& & &  Medicine& Tasks& Engineering&  \\ \hline
\textbf{Human Evaluation} & \textbf{31.183} & \textbf{59.782} & \textbf{42.103} & \textbf{48.858} & \textbf{56.429} &\textbf{50.134} & \textbf{43.758}\\
\hdashline
GPT-4V~\citep{gpt4-v}         & 30.399           & 89.203           & 68.731           & 80.059           & 38.432           & 69.108   &48.793             \\
Gemini-Pro~\citep{team2023gemini}          & 28.745           & 80.909           & 69.425           & 80.023           & 50.987           & 80.479      & 48.083                \\
\bottomrule
\end{tabular}}
\label{tab:subset_human_eval}
\end{table}

We also hired students from campus to do human evaluation on subset of the data. The results are shown in Table~\ref{tab:subset_human_eval}. The performance of the human evaluators did not surpass that of GPT-4V and Gemini-Pro. This outcome underscores the challenging nature of the dataset, which often necessitates specialized domain knowledge that our evaluators—primarily non-experts—found demanding. These results highlight the complexity of the questions and the potential necessity for discipline-specific understanding to achieve high accuracy

\subsection{Quality of Using GPT as the Judger}
For a comprehensive assessment of GPT-4V's accuracy when using it as the judger, we devised a human evaluation protocol also resort to Amazon
Mechanical Turk, as visualized in Figure~\ref{fig:human_eval}. The evaluators present a series of statements derived from the video, and GPT-4V is tasked with selecting the most accurate answer from a set of multiple-choice questions. Through this interface, human evaluators can efficiently gauge GPT-4V's performance across different types of questions—when using it as the judger.

The results obtained from this human evaluation process are shown in Table~\ref{tab:error_rates}, across 189 examples, there are only 9 incorrect ones with the error rate of 4.76\%, validating the effectiveness of using GPT-4V as the judger.

\begin{figure}[tb]
  \centering
  \includegraphics[width=\textwidth]{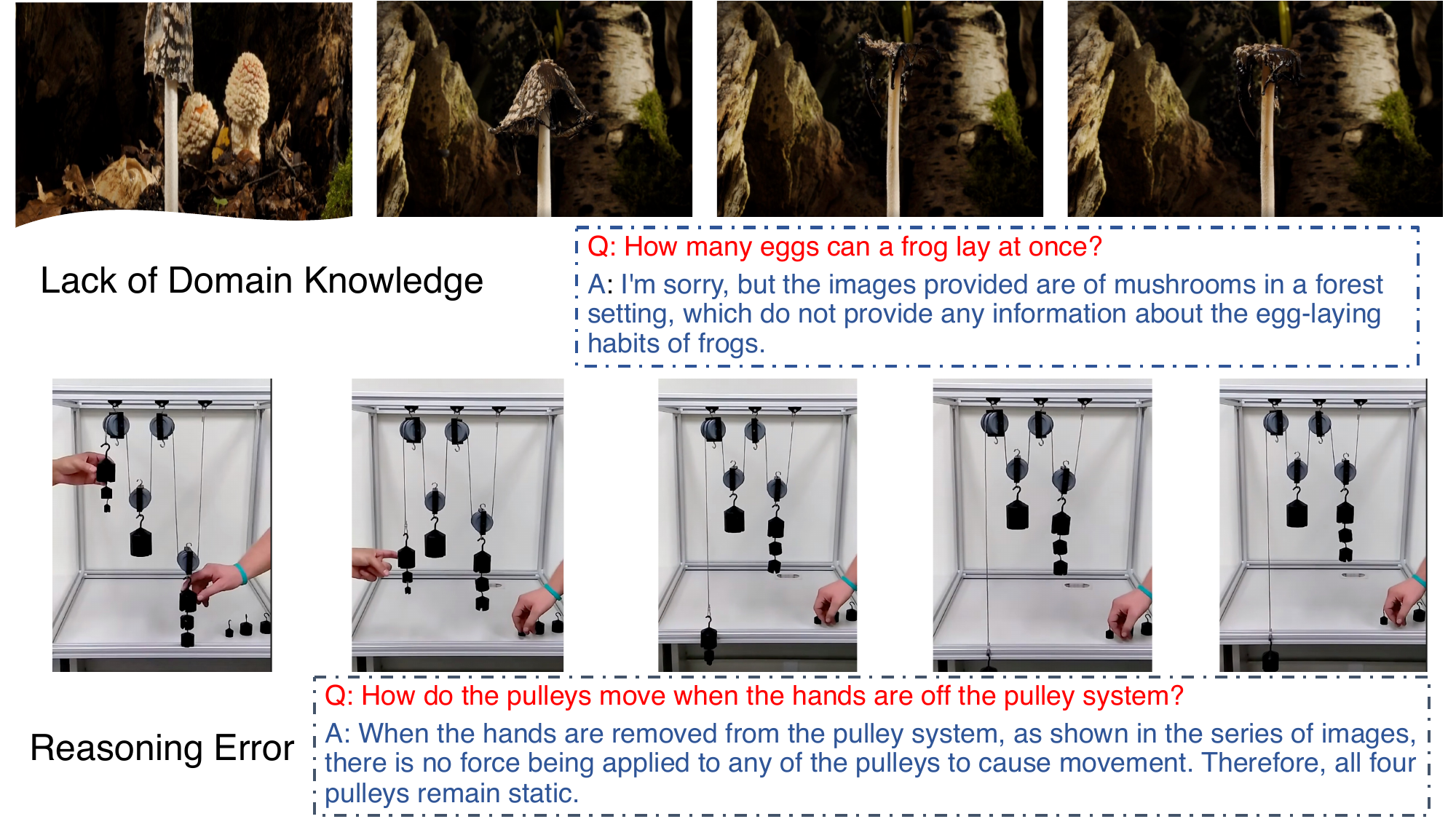}
  \caption{Error pattern of lack of domain knowledge and reasoning error. In the first case, the model does not give the correct answer because of lacking the domain knowledge. In the second case, the model makes the wrong reasoning.
  }
  \label{fig:error1}
\end{figure}

\begin{figure}[tb]
  \centering
  \includegraphics[width=\textwidth]{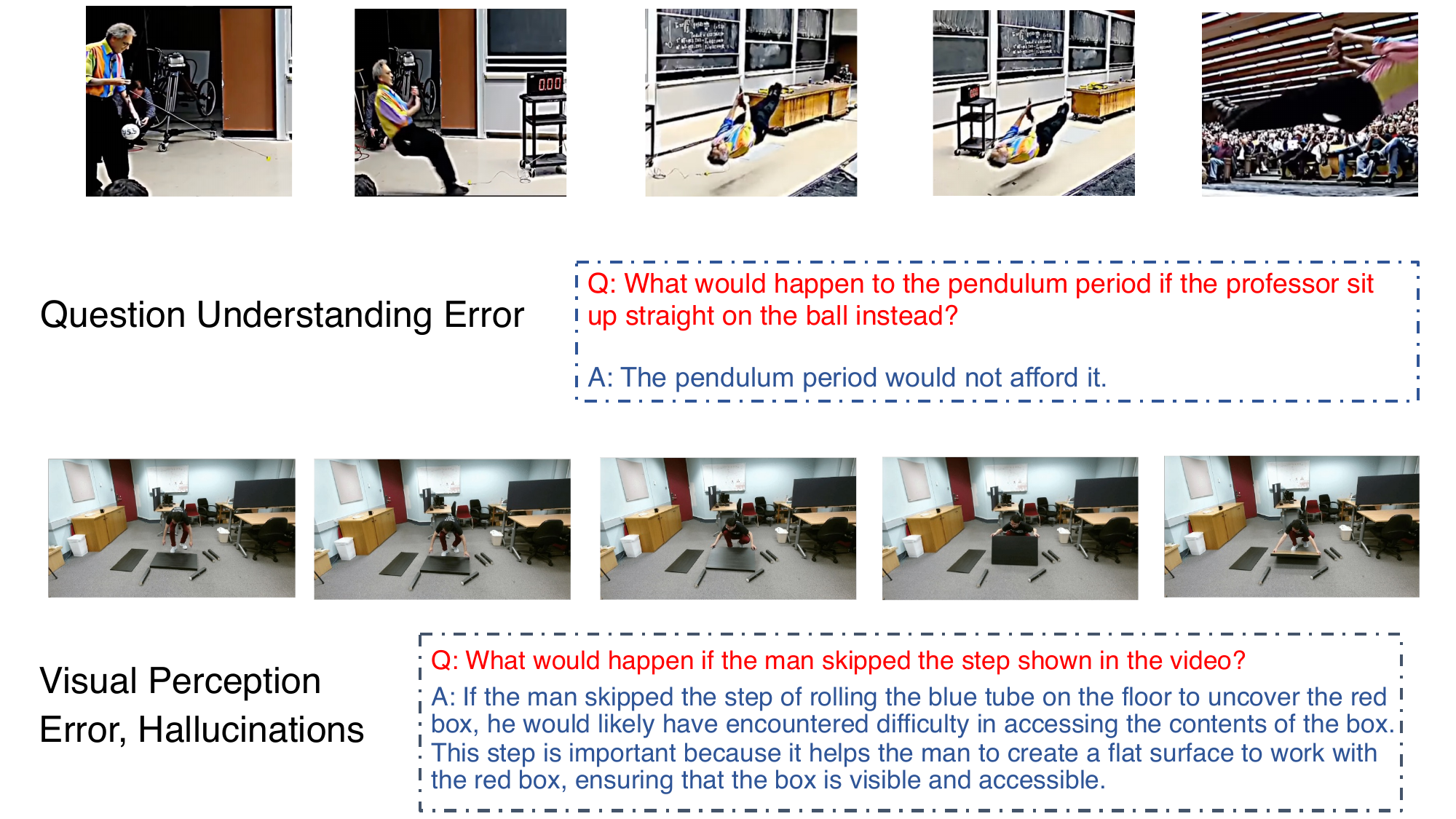}
  \caption{Error pattern of question understanding, visual perceptron error, and hallucinations. In the first case, the model does not understand the question correctly where the model fails to accurately discern the query regarding the pendulum's period. In the second scenario, the model erroneously identifies objects within the visual input, leading to the hallucination of non-existent elements, such as a red box.
  }
  \label{fig:error2}
\end{figure}

\begin{figure}[tb]
  \centering
  \includegraphics[width=\textwidth]{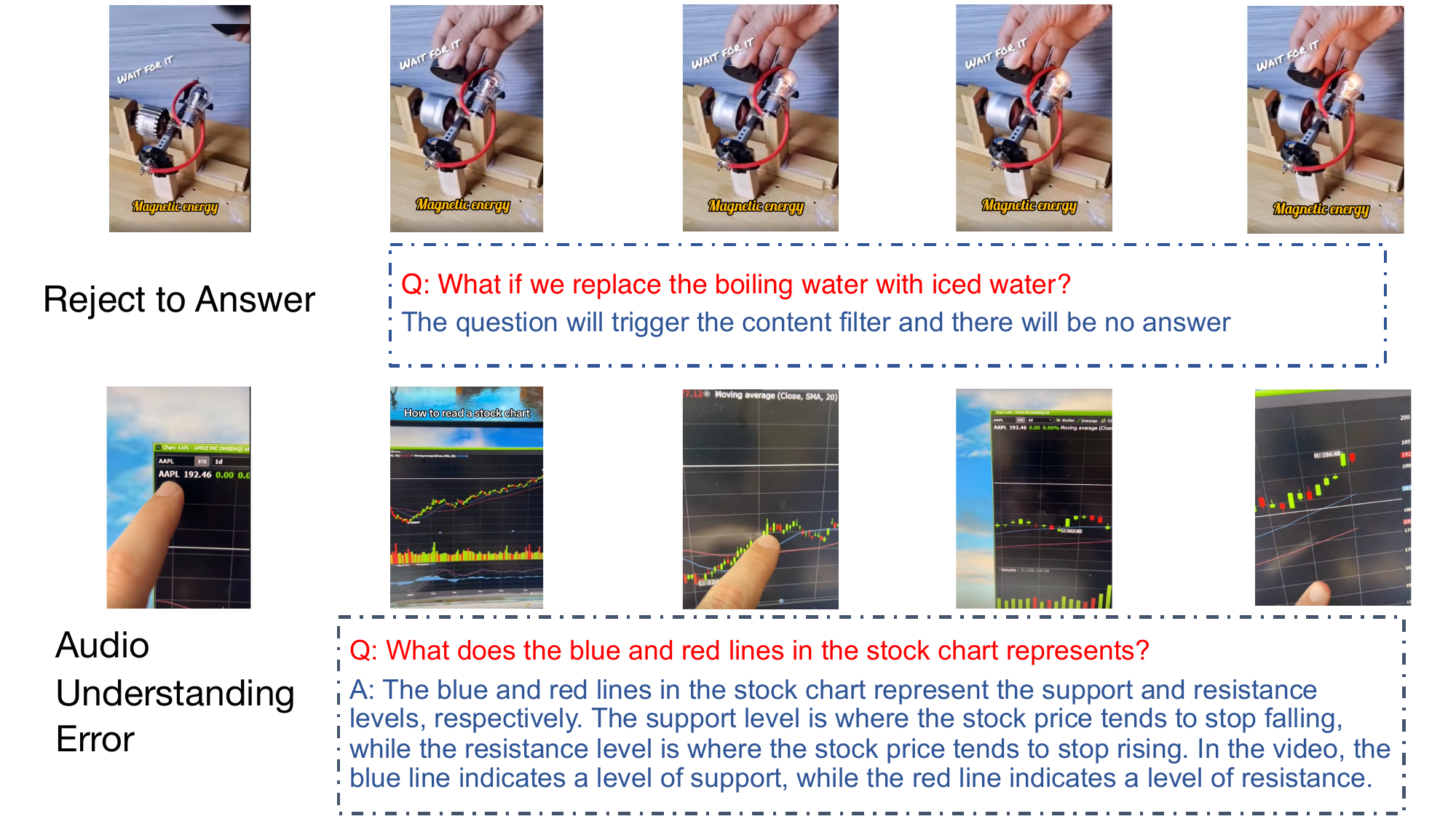}
  \caption{Error pattern of reject to answer and audio understanding error. In the first case, the model will reject to answer because of the dangerous action showing in the video. In the second case, the model does not understand the audio correctly as the answer to the question is given in the audio of the video.
  }
  \label{fig:error3}
\end{figure}

\begin{figure}[tb]
  \centering
  \includegraphics[width=\textwidth]{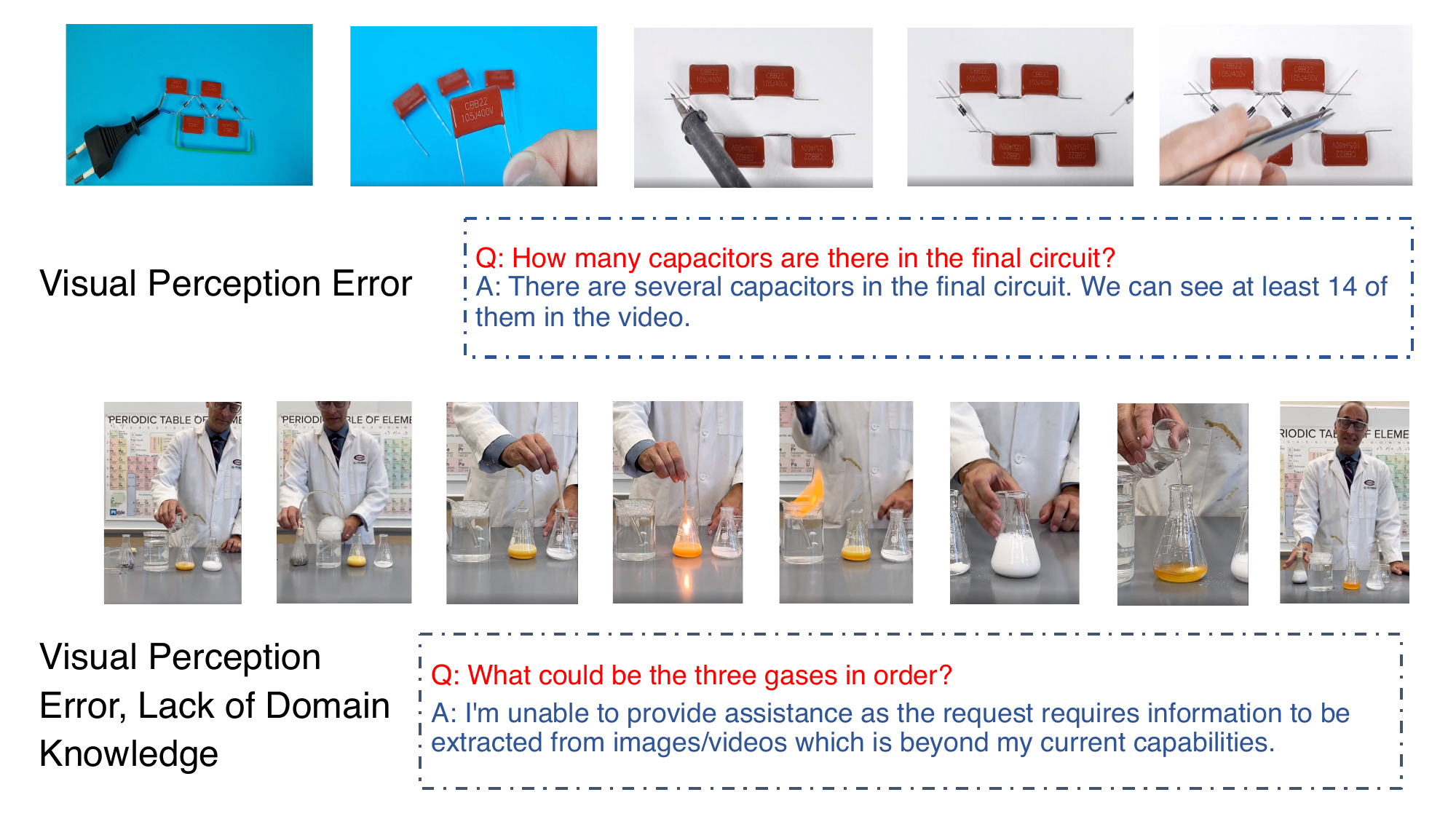}
  \caption{Error pattern due to visual perception inaccuracies and insufficient domain knowledge. The first case demonstrates a visual perception error where the model incorrectly identifies the number of capacitors present. The second case showcases a compound error where the model not only fails to discern the colors indicative of different gases but also lacks the domain knowledge necessary to infer their identity correctly.
  }
  \label{fig:error4}
\end{figure}

\begin{figure}[t]
    \centering
    \includegraphics[width=\textwidth]{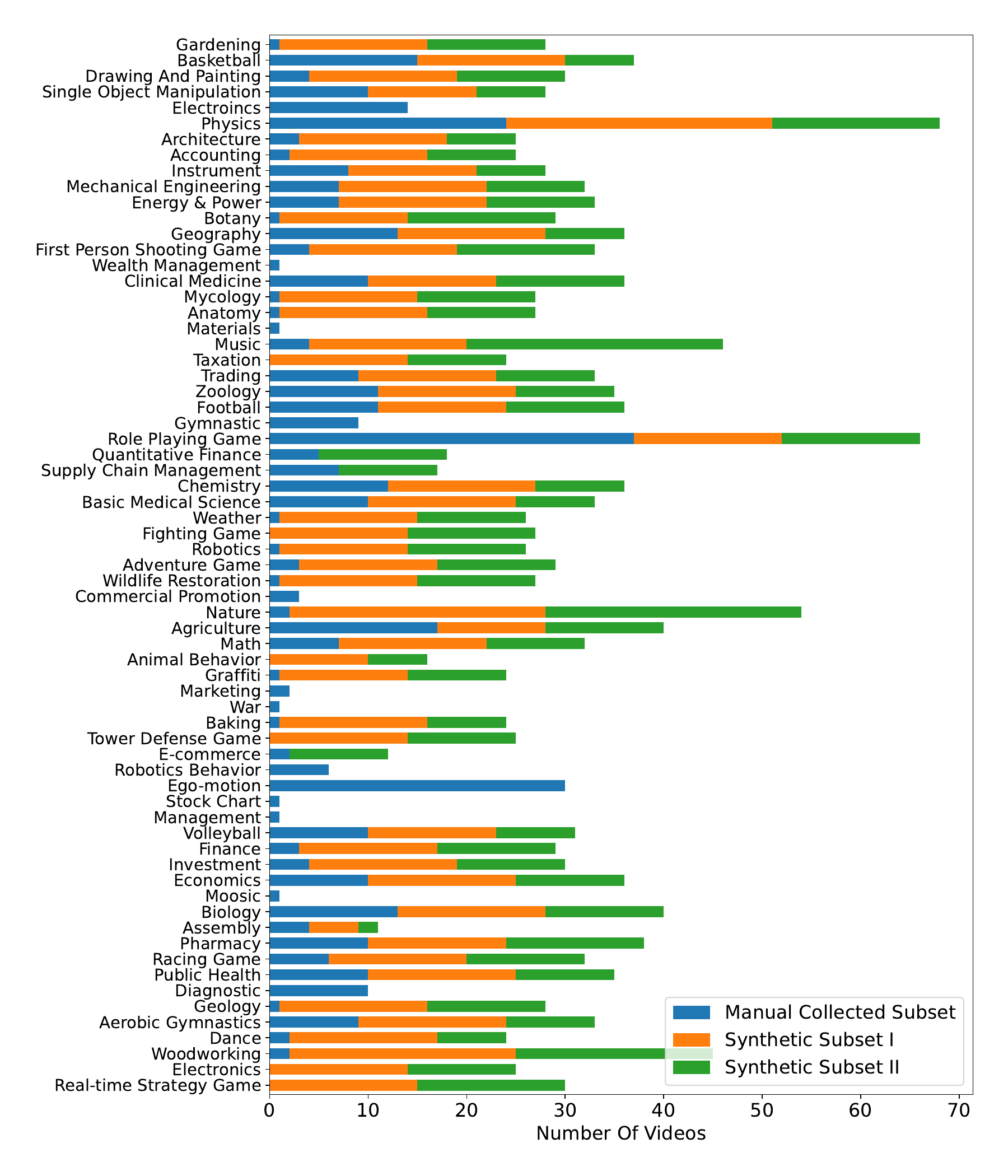}
    \caption{The number of videos per subdiscipline in \benchmarkname. Each horizontal bar indicates the quantity of videos corresponding to a subdiscipline, showcasing the dataset's diversity and coverage across various domains of knowledge. Synthetic Subset I is collected with audio-only data and Synthetic Subset II is collected with visual-only data. 
    }
    \label{fig:subdomain}
\end{figure}

\begin{figure}[tb]
  \centering
  \includegraphics[width=0.6\textwidth]{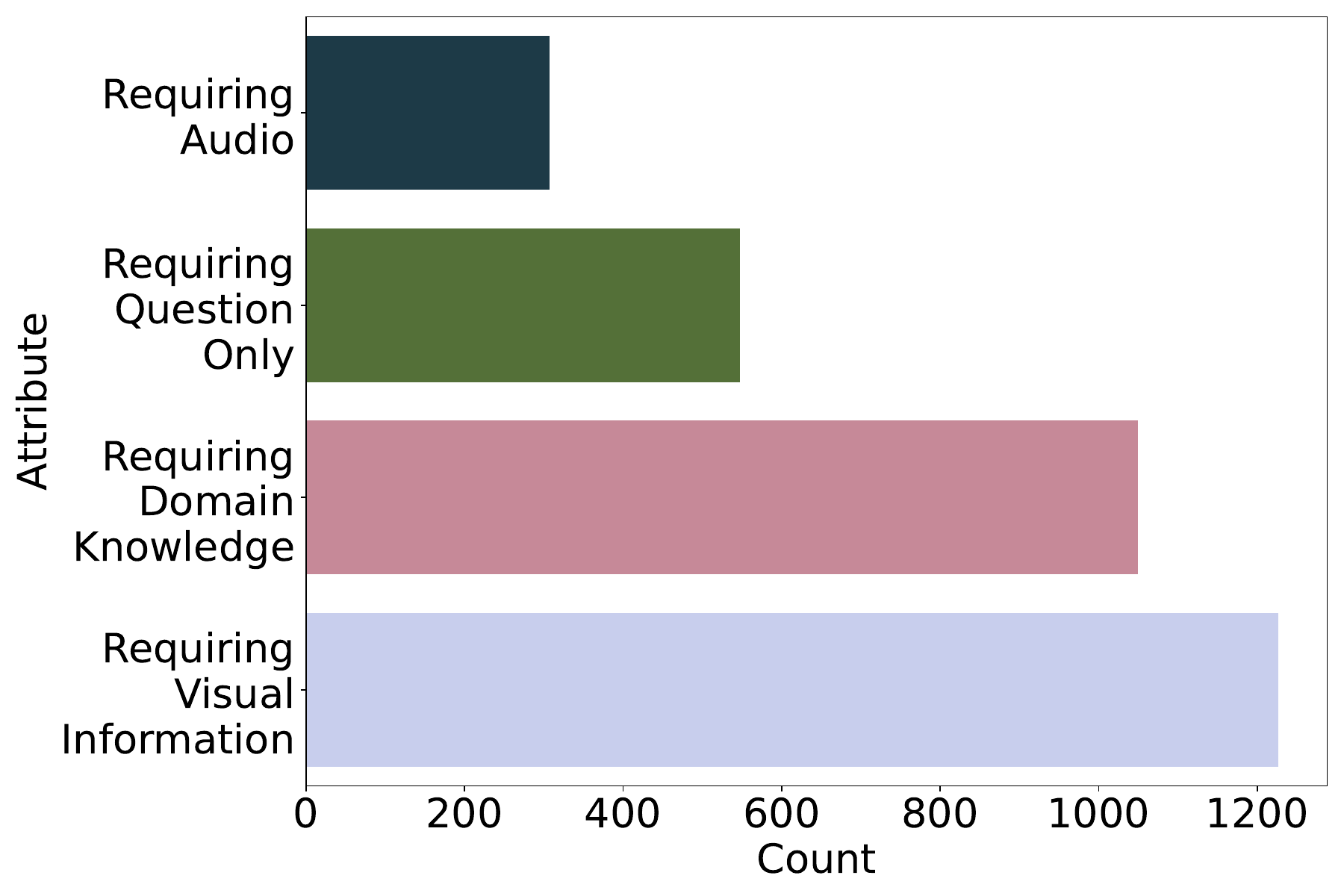}
  \caption{The distribution statistics of questions in the \benchmarkname benchmark by annotations. }
  \label{fig:questions_per_annotation}
\end{figure}

\section{Error Analysis}
\label{error}
In this section, we delve into the analysis of errors from evaluated MLLMs. We summarized error types as follows:

\MyParagraph{Question Understanding Error (QUE):} Models misinterpret the question's intent, such as misunderstanding how a pendulum's period would change if a condition in the scenario is altered.

\MyParagraph{Audio Understanding Error (AUE):} Models fail to interpret audio cues correctly, shown by their failure to recognize blue and red lines on a stock chart. 

\MyParagraph{Visual Perception Error (VPE):} There is a misinterpretation of visual content, leading to incorrect assumptions about the visual data presented in the video.

\MyParagraph{Hallucinations (HE):} Models generate content or details that are not present in the actual data, essentially `hallucinating' information.

\MyParagraph{Reasoning Error (RE):} Models demonstrate a lack of logical reasoning, leading to incorrect conclusions based on the given data.

\MyParagraph{Lack of Domain Knowledge (LDK):} Models show an inability to answer questions that require specific domain expertise, indicating a gap in their knowledge.

\MyParagraph{Reject to Answer (RA):} An example of this error was observed when the model was asked to select an answer regarding the outcome of an experiment involving liquid nitrogen. Instead of choosing an option, the model provided an unrelated response concerning a light bulb, indicating either a misunderstanding or a cautious approach due to the potential for the question to be interpreted as pertaining to a sensitive topic, which can trigger content filters focused on safety and compliance policies.

We show in Figure~\ref{fig:error1},~\ref{fig:error2},~\ref{fig:error3},~\ref{fig:error4} some error cases of \textit{Question Understanding Error}, \textit{Audio Understanding Error}, \textit{Visual Perception Error}, \textit{Hallucinations}, \textit{Reasoning Error}, \textit{Lack of Domain Knowledge}, and \textit{Reject to Answer} respectively from MLLMs evaluated on \benchmarkname.

\section{Data Examples}
\label{data}
We show in Figure~\ref{fig:data_examples1},~\ref{fig:data_examples2},~\ref{fig:data_examples3},~\ref{fig:data_examples4},~\ref{fig:data_examples5},~\ref{fig:temp_example} some additional examples from \benchmarkname.

\section{Additional Data Statistics}
\label{subdiscipline_statistics}
For human annotated dataset, the length of each video was capped at approximately two minutes. The statistical distribution of the disciplines within the dataset for this part is as follows:
\begin{itemize}
 \item  \emph{Sports \& Arts}: The subset that consists of 77 videos, showcasing a vibrant collection that covers a wide range of topics from athletic endeavors to various forms of artistic expression.
 \item  \emph{Science}: A subset of 75 videos, which delves into the empirical world of scientific inquiry, spanning a multitude of specializations from fundamental physics to advanced biological studies.
 \item  \emph{Tech \& Engineering}: Encompassing 54 videos, this segment captures the cutting-edge advancements and foundational concepts that drive innovation and infrastructure in the modern world.
 \item  \emph{Embodied Tasks}: With 50 videos, the dataset provides a focused insight into the dynamic field of Embodied Tasks, highlighting the intersection of AI, mechanics, and automation.
 \item  \emph{Health \& Medicine}: This essential discipline is well-represented with 50 videos, offering perspectives on medical breakthroughs, healthcare practices, and life sciences.
 \item  \emph{Business}: This discipline includes 50 videos, reflecting on the multifaceted nature of commerce, from economics to management sciences.
  \item  \emph{Game}: This discipline includes 51 videos, reflecting various aspects of gaming.
\end{itemize}

Altogether, the \benchmarkname Benchmark's diversity is visually encapsulated in Figure~\ref{fig:subdomain}, which delineates the distribution of videos across 61 subdisciplines. The horizontal bar chart provides a quantified representation of the dataset's range, reflecting the careful curation process that has gone into ensuring breadth across various knowledge areas.

The world we live in is rich with both audio and visual information, and effective world modeling requires an understanding of how these modalities interact and convey meaning. To achieve this, we annotated additional attributes such as "Requires Audio," "Requires Video," and "Question Only." These annotations help determine whether correctly answering a question necessitates audio information, visual cues from the video, or can be addressed based solely on the question itself. By doing so, we ensure that our benchmark tests the full spectrum of multimodal comprehension, reflecting the complex, sensory-rich environment in which real-world understanding takes place. The statistics of these annotations are shown in Figure~\ref{fig:questions_per_annotation}.

\section{Datasheets}
\label{sec:datasheet}
\subsection{Motivation}
\textbf{For what purpose was the dataset created?} 

To introduce a multi-discipline multi-faceted multimodal video understanding benchmark to comprehensively evaluate MLLMs’ abilities in reasoning and interpreting real-world dynamics.

\textbf{Who created the dataset (e.g., which team, research group) and on behalf of which entity (e.g., company, institution, organization)?} 

The dataset is created by authors from UCSC, UCSB, and Microsoft.

\textbf{Who funded the creation of the dataset?} 

UCSC, UCSB, and Microsoft Azure.

\subsection{Composition}
\textbf{What do the instances that comprise the dataset represent? (e.g., documents, photos, people, countries)} 

Videos along with captions and question/answer pairs.

\textbf{How many instances are there in total (of each type, if appropriate)?} 

6,627 instances. The data distribution over different types can be found in Figure 2 of the main paper.

\textbf{Does the dataset contain all possible instances or is it a sample (not necessarily random) of instances from a larger set?} 

Yes.

\textbf{Is there a label or target associated with each instance?} 

Yes.

\textbf{Is any information missing from individual instances?} 

No.

\textbf{Are relationships between individual instances made explicit (e.g., users’ movie ratings, social network links)?} 

N/A.

\textbf{Are there recommended data splits (e.g., training, development/validation, testing)?}  

The \benchmarkname is used for evaluation purpose only.

\textbf{Are there any errors, sources of noise, or redundancies in the dataset?} 

No.

\textbf{Is the dataset self-contained, or does it link to or otherwise rely on external resources (e.g., websites, tweets, other datasets)?}

Yes.

\textbf{Does the dataset contain data that might be considered confidential?} 

No.

\textbf{Does the dataset contain data that, if viewed directly, might be offensive, insulting, threatening, or might otherwise cause anxiety?} 

No.

\subsection{Collection Process}
The data collection process is described in Section 3 of the main paper.

\subsection{Preprocessing/cleaning/labeling
}
\textbf{Was any preprocessing/cleaning/labeling of the data done (e.g., discretization or bucketing, tokenization, part-of-speech tagging, SIFT feature extraction, removal of instances, processing of missing values}

We extract video frames from collected videos in automatically generated.

\textbf{Was the “raw” data saved in addition to the preprocessed/cleaned/labeled data (e.g., to support unanticipated future uses)?}

Yes. The raw video urls are given.

\textbf{Is the software that was used to preprocess/clean/label the data
available?}

Yes. The source code can be found in \url{https://github.com/eric-ai-lab/MMWorld}.

\subsection{Uses}
\textbf{Has the dataset been used for any tasks already?} 

Yes. We have used the dataset to evaluate video question answering.

\textbf{Is there a repository that links to any or all papers or systems that use the dataset?} 

Yes. The GitHub repository \url{https://github.com/eric-ai-lab/MMWorld} here.

\textbf{What (other) tasks could the dataset be used for?} 

Video captioning and evaluating faithfulness of evaluation metrics.

\textbf{Is there anything about the composition of the dataset or the way it was collected and preprocessed/cleaned/labeled that might impact future uses?} 

No.

\textbf{Are there tasks for which the dataset should not be used?} 

The videos in this dataset are from different sources and are unique. The dataset should not be used for tasks such as video editing.

\subsection{Distribution}
\textbf{Will the dataset be distributed to third parties outside of the entity (e.g., company, institution, organization) on behalf of which the dataset was created?} 

Yes. The benchmark is publicly available.

\textbf{How will the dataset will be distributed (e.g., tarball on website, API, GitHub)?} 

We host it on the webpage, GitHub, and Huggingface.

\textbf{When will the dataset be distributed?}

It's availale and open to the public now.

\textbf{Will the dataset be distributed under a copyright or other intellectual property (IP) license, and/or under applicable terms of use (ToU)?} 

CC-By 4.0.

\textbf{Have any third parties imposed IP-based or other restrictions on the data associated with the instances?} 

No.

\textbf{Do any export controls or other regulatory restrictions apply to the dataset or to individual instances?} 

No.

\subsection{Maintenance}
\textbf{Who will be supporting/hosting/maintaining the dataset?} 

The authors will be supporting/hosting/maintaining the dataset.

\textbf{How can the owner/curator/manager of the dataset be contacted (e.g., email address)?} 

The email address is xhe89@ucsc.edu.

\textbf{Is there an erratum?} 

No. We will make it if there is any erratum.

\textbf{Will the dataset be updated (e.g., to correct labeling errors, add new instances, delete instances)?} 

Yes. We will make announcements on GitHub if there is any update.

\textbf{If the dataset relates to people, are there applicable limits on the retention of the data associated with the instances (e.g., were individuals in question told that their data would be retained for a fixed period of time and then deleted)?} 

N/A.

\textbf{Will older versions of the dataset continue to be supported/hosted/maintained?} 

Yes. Old versions can still be accessed from Huggingface.

\textbf{If others want to extend/augment/build on/contribute to the dataset, is there a mechanism for them to do so?} 

Yes. Contributors can post issues or submit pull requests on GitHub. We will review and verify contributions, and update the dataset if the contribution is useful.

\section{Author Statement, Hosting, Licensing, and Maintenance Plan}
\label{sec:license_host}

\paragraph{Author Statement}

We bear all responsibility in case of violation of rights and confirmation of the data license.

\paragraph{Hosting}

\benchmarkname is hosted on \url{https://mmworld-bench.github.io/}. The dataset is provided in the JSON file format. The metadata can be found at \url{https://huggingface.co/datasets/Xuehai/MMWorld}.

\paragraph{License}

\benchmarkname is licensed under the CC-BY 4.0 license.

\paragraph{Maintenance Plan}

We will keep maintaining and updating the dataset and benchmark, including the leaderboard.





\end{document}